\documentclass{article}
\usepackage{log_2026}
\makeatletter
\renewcommand{\@noticestring}{Accepted at the Fifth Learning on Graphs Conference (LoG 2026).}
\makeatother
\AtEndPreamble{\usepackage[capitalise,nameinlink,noabbrev]{cleveref}}

\usepackage{amssymb}
\usepackage{amsfonts}
\usepackage{array}
\usepackage{booktabs}
\usepackage{float}
\usepackage{graphicx}
\usepackage{pgfplots}
\usepgfplotslibrary{groupplots}
\usetikzlibrary{patterns}
\pgfplotsset{compat=1.18}
\usepackage{multirow}
\usepackage[numbers,compress,sort]{natbib}
\usepackage{adjustbox}
\usepackage{threeparttable}
\usepackage{arydshln}
\usepackage{iftex}

\ifPDFTeX
\usepackage{fontawesome5}
\else
\newcommand{\faIcon}[1]{%
    \ifstrequal{#1}{hashtag}{\textsf{\#}}{%
    \ifstrequal{#1}{share-alt}{\textsf{L}}{%
    \ifstrequal{#1}{globe}{\textsf{G}}{%
    \ifstrequal{#1}{cogs}{\textsf{S}}{?}}}}}
\newcommand{\faClone}{\textsf{C}}
\newcommand{\faLink}{\textsf{L}}
\newcommand{\faCogs}{\textsf{D}}
\newcommand{\faSortNumericDown}{\textsf{R}}
\fi

\newcommand{\evOpaque}{\faIcon{hashtag}\xspace}
\newcommand{\evLocal}{\faIcon{share-alt}\xspace}
\newcommand{\evGlobal}{\faIcon{globe}\xspace}
\newcommand{\evScore}{\faIcon{cogs}\xspace}

\newcommand{\evLocalGlobal}{%
    \faIcon{share-alt}\faIcon{globe}\xspace}

\newcommand{\evLocalScore}{%
    \faIcon{share-alt}\faIcon{cogs}\xspace}

\newcommand{\evFull}{%
    \faIcon{share-alt}\faIcon{cogs}\faIcon{globe}\xspace}

\newcommand{\ClutrrLogMultiset}{0.429}
\newcommand{\ClutrrCatMultiset}{0.272}
\newcommand{\ClutrrTabMultiset}{0.289}
\newcommand{\ClutrrLogAbsolute}{0.156}
\newcommand{\ClutrrCatAbsolute}{0.109}
\newcommand{\ClutrrTabAbsolute}{0.103}
\newcommand{\ClutrrLogCombined}{0.266}
\newcommand{\ClutrrCatCombined}{0.163}
\newcommand{\ClutrrTabCombined}{0.187}
\newcommand{\ClutrrLogPermuted}{0.399}
\newcommand{\ClutrrCatPermuted}{0.255}
\newcommand{\ClutrrTabPermuted}{0.291}
\newcommand{\ClutrrLogTransition}{0.408}
\newcommand{\ClutrrCatTransition}{0.285}
\newcommand{\ClutrrTabTransition}{0.156}
\newcommand{\ClutrrLogAlignCost}{0.133}
\newcommand{\ClutrrCatAlignCost}{0.092}
\newcommand{\ClutrrTabAlignCost}{0.104}
\newcommand{\ClutrrAbsoluteCeiling}{1.000}
\newcommand{\ClutrrAbsoluteSingleton}{1.000}
\newcommand{\ClutrrMultisetCeiling}{0.930}
\newcommand{\ClutrrMultisetSingleton}{0.757}
\newcommand{\ClutrrMultisetNonSingletonMass}{0.243}
\newcommand{\ClutrrMultisetNonSingletonCeiling}{0.712}
\newcommand{\ClutrrPrefixCeiling}{0.998}
\newcommand{\ClutrrPrefixSingleton}{0.991}
\newcommand{\ClutrrPrefixNonSingletonMass}{0.009}
\newcommand{\ClutrrPrefixNonSingletonCeiling}{0.721}
\newcommand{\ClutrrContextFitRange}{0.990--0.995}
\newcommand{\ClutrrAddLogFour}{+0.489}
\newcommand{\ClutrrAddCatFour}{+0.372}
\newcommand{\ClutrrAddTabFour}{+0.758}
\newcommand{\ClutrrAddLogLong}{-0.001}
\newcommand{\ClutrrAddCatLong}{+0.010}
\newcommand{\ClutrrAddTabLong}{+0.033}
\newcommand{\ClutrrGru}{0.705}
\newcommand{\ClutrrGruContextLow}{0.346}
\newcommand{\ClutrrGruContextHigh}{0.705}
\newcommand{\ClutrrGruLengthFour}{0.898}
\newcommand{\ClutrrGruLengthTen}{0.527}

\newcommand{\GraphLogLogShared}{0.298}
\newcommand{\GraphLogLogAbsolute}{0.234}
\newcommand{\GraphLogCatShared}{0.125}
\newcommand{\GraphLogCatAbsolute}{0.282}
\newcommand{\GraphLogTabShared}{0.159}
\newcommand{\GraphLogTabAbsolute}{0.238}
\newcommand{\GraphLogLogDelta}{+0.065}
\newcommand{\GraphLogCatDelta}{-0.157}
\newcommand{\GraphLogTabDelta}{-0.079}

\newcommand{\GraphLogSharedCeiling}{0.965}
\newcommand{\GraphLogAbsoluteCeiling}{0.983}
\newcommand{\GraphLogDepthOneDelta}{-0.070}
\newcommand{\GraphLogDepthSevenDelta}{-0.157}

\newcommand{\GraphLogDepthOneShared}{0.201}
\newcommand{\GraphLogDepthOneAbsolute}{0.271}
\newcommand{\GraphLogDepthSevenShared}{0.125}
\newcommand{\GraphLogDepthSevenAbsolute}{0.282}
\newcommand{\GraphLogDepthOneContextShared}{0.830}
\newcommand{\GraphLogDepthOneContextAbsolute}{0.910}
\newcommand{\GraphLogDepthSevenContextShared}{0.948}
\newcommand{\GraphLogDepthSevenContextAbsolute}{0.991}

\newcommand{\DEightChance}{0.0625}
\newcommand{\DEightLogShared}{0.0618}
\newcommand{\DEightLogAbsolute}{0.0622}
\newcommand{\DEightCatShared}{0.0770}
\newcommand{\DEightCatAbsolute}{0.0643}
\newcommand{\DEightTabShared}{0.0614}
\newcommand{\DEightTabAbsolute}{0.0626}

\newcommand{\DEightLogTransition}{0.0622}
\newcommand{\DEightCatTransition}{0.0617}
\newcommand{\DEightTabTransition}{0.0648}

\newcommand{\UltraDatasets}{57}
\newcommand{\UltraFamilies}{16}

\newcommand{\UltraGlobalGivenScore}{+0.005}
\newcommand{\UltraGlobalGivenScoreCI}{[-0.001,+0.010]}
\newcommand{\UltraFullNative}{+0.006}
\newcommand{\UltraFullNativeCI}{[-0.053,+0.108]}
\newcommand{\UltraFullNativeWTL}{5/4/48}
\newcommand{\UltraFullNativeNoMeta}{-0.041}
\newcommand{\UltraFullNativeFamilyMedian}{-0.034}
\newcommand{\UltraFullNativeDatasetMean}{-0.035}
\newcommand{\UltraFullNativeDatasetMedian}{-0.039}
\newcommand{\UltraFullNativeLOFO}{[-0.041,+0.015]}

\newcommand{\UltraScoreRankAuditCells}{684}

\newcommand{\ContextLocalIdsRange}{+0.177--+0.262}
\newcommand{\ContextScoreLocalRange}{+0.048--+0.129}

\newcommand{\ContextInformativeDropRange}{0.020--0.122}

\newcommand{\ContextLogLocalUniform}{0.199}
\newcommand{\ContextCatLocalUniform}{0.262}
\newcommand{\ContextTabLocalUniform}{0.267}
\newcommand{\ContextLogScoreUniform}{0.335}
\newcommand{\ContextCatScoreUniform}{0.299}
\newcommand{\ContextTabScoreUniform}{0.332}
\newcommand{\ContextLogLocalScoreUniformLevel}{0.328}
\newcommand{\ContextCatLocalScoreUniformLevel}{0.310}
\newcommand{\ContextTabLocalScoreUniformLevel}{0.391}
\newcommand{\ContextLogLocalStructural}{0.179}
\newcommand{\ContextCatLocalStructural}{0.199}
\newcommand{\ContextTabLocalStructural}{0.213}
\newcommand{\ContextLogScoreStructural}{0.301}
\newcommand{\ContextCatScoreStructural}{0.197}
\newcommand{\ContextTabScoreStructural}{0.245}
\newcommand{\ContextLogLocalScoreStructuralLevel}{0.262}
\newcommand{\ContextCatLocalScoreStructuralLevel}{0.261}
\newcommand{\ContextTabLocalScoreStructuralLevel}{0.269}

\newcommand{\QueryRelLogLocalRaw}{0.198}
\newcommand{\QueryRelLogLocalZ}{0.279}
\newcommand{\QueryRelLogLocalPct}{0.274}
\newcommand{\QueryRelLogLocalZDelta}{+0.081}
\newcommand{\QueryRelLogLocalPctDelta}{+0.076}
\newcommand{\QueryRelLogLocalScoreRaw}{0.326}
\newcommand{\QueryRelLogLocalScoreZ}{0.285}
\newcommand{\QueryRelLogLocalScorePct}{0.274}
\newcommand{\QueryRelLogLocalScoreZDelta}{-0.041}
\newcommand{\QueryRelLogLocalScorePctDelta}{-0.053}
\newcommand{\QueryRelLogScoreRaw}{0.335}
\newcommand{\QueryRelLogScoreZ}{0.335}
\newcommand{\QueryRelLogScorePct}{0.335}
\newcommand{\QueryRelCatLocalRaw}{0.245}
\newcommand{\QueryRelCatLocalZ}{0.199}
\newcommand{\QueryRelCatLocalPct}{0.181}
\newcommand{\QueryRelCatLocalZDelta}{-0.046}
\newcommand{\QueryRelCatLocalPctDelta}{-0.064}
\newcommand{\QueryRelCatLocalScoreRaw}{0.301}
\newcommand{\QueryRelCatLocalScoreZ}{0.159}
\newcommand{\QueryRelCatLocalScorePct}{0.145}
\newcommand{\QueryRelCatLocalScoreZDelta}{-0.142}
\newcommand{\QueryRelCatLocalScorePctDelta}{-0.156}
\newcommand{\QueryRelCatScoreRaw}{0.299}
\newcommand{\QueryRelCatScoreZ}{0.006}
\newcommand{\QueryRelCatScorePct}{0.004}
\newcommand{\QueryRelCatScoreZDelta}{-0.293}
\newcommand{\QueryRelCatScorePctDelta}{-0.295}
\newcommand{\QueryRelTabLocalRaw}{0.265}
\newcommand{\QueryRelTabLocalZ}{0.002}
\newcommand{\QueryRelTabLocalPct}{0.142}
\newcommand{\QueryRelTabLocalZDelta}{-0.262}
\newcommand{\QueryRelTabLocalPctDelta}{-0.123}
\newcommand{\QueryRelTabLocalScoreRaw}{0.387}
\newcommand{\QueryRelTabLocalScoreZ}{0.002}
\newcommand{\QueryRelTabLocalScorePct}{0.222}
\newcommand{\QueryRelTabLocalScoreZDelta}{-0.384}
\newcommand{\QueryRelTabLocalScorePctDelta}{-0.165}
\newcommand{\QueryRelTabScoreRaw}{0.329}
\newcommand{\QueryRelTabScoreZ}{0.002}
\newcommand{\QueryRelTabScorePct}{0.186}
\newcommand{\QueryRelTabScoreZDelta}{-0.326}
\newcommand{\QueryRelTabScorePctDelta}{-0.142}
\newcommand{\QueryRelCells}{4,617}
\newcommand{\QueryRelRankCells}{171}
\newcommand{\QueryRelRanks}{21,888}

\newcommand{\FilterLocalUnfiltered}{0.257}
\newcommand{\FilterLocalFiltered}{0.265}
\newcommand{\FilterLocalFamilyDelta}{+0.0081}
\newcommand{\FilterLocalDatasetDelta}{+0.0056}
\newcommand{\FilterLocalSensitivity}{[+0.0022,+0.0147]}
\newcommand{\FilterLocalMaxP}{0.0578}
\newcommand{\FilterLocalScoreUnfiltered}{0.387}
\newcommand{\FilterLocalScoreFiltered}{0.387}
\newcommand{\FilterLocalScoreFamilyDelta}{-0.0007}
\newcommand{\FilterLocalScoreDatasetDelta}{+0.0040}
\newcommand{\FilterLocalScoreSensitivity}{[-0.0085,+0.0062]}
\newcommand{\FilterLocalScoreMaxP}{0.9972}
\newcommand{\FilterScoreUnfiltered}{0.326}
\newcommand{\FilterScoreFiltered}{0.329}
\newcommand{\FilterScoreFamilyDelta}{+0.0022}
\newcommand{\FilterScoreDatasetDelta}{+0.0021}
\newcommand{\FilterScoreSensitivity}{[-0.0022,+0.0074]}
\newcommand{\FilterScoreMaxP}{0.7739}
\newcommand{\FilterCells}{1,026}

\newcommand{\QueryBoth}{244}

\newcommand{\QueryDirect}{414}
\newcommand{\QueryNeither}{5,674}

\newcommand{\QueryDyadicNeitherRange}{0.001--0.002}
\newcommand{\QueryPathsNeitherRange}{0.045--0.052}
\newcommand{\QueryLocalNeitherRange}{0.043--0.094}

\newcommand{\ScreenRankSpearman}{0.957}
\newcommand{\ScreenStrata}{455}
\newcommand{\ScreenAssociationPairs}{56}

\newcommand{\QueryVarianceShare}{0.250}
\newcommand{\InterfaceVarianceShare}{0.147}
\newcommand{\DecoderVarianceShare}{0.003}
\newcommand{\QueryDecoderInteractionShare}{0.028}
\newcommand{\QueryTargetDegreeRho}{+0.51\text{ to }+0.58}
\newcommand{\QueryAnchorDegreeRho}{-0.32\text{ to }-0.23}

\newcommand{\ControlRelationCeiling}{0.520}
\newcommand{\ControlRelationCollision}{1.000}
\newcommand{\ControlReciprocityCeiling}{1.000}
\newcommand{\ControlReciprocitySingleton}{1.000}
\newcommand{\ControlCompositionCeiling}{1.000}
\newcommand{\ControlCompositionSingleton}{0.067}
\newcommand{\ControlRolesCeiling}{1.000}
\newcommand{\ControlRolesSingleton}{0.000}
\newcommand{\ControlBlockCeiling}{0.924}
\newcommand{\ControlBlockCollision}{0.191}

\newcommand{\TabPFN}{\textsc{TabPFN}\xspace}
\newcommand{\CatBoost}{\textsc{CatBoost}\xspace}
\newcommand{\Ultra}{\textsc{Ultra}\xspace}
\newcommand{\TabPFNthree}{\textsc{TabPFN-3}\xspace}

\newcommand{\Astar}{A^\star\xspace}

\newcommand{\context}{\mathcal{S}\xspace}

\definecolor{gttEncoderFill}{RGB}{116,174,210}
\definecolor{gttContextFill}{RGB}{236,190,103}
\definecolor{gttDecoderFill}{RGB}{108,184,168}

\definecolor{gttEncoderText}{RGB}{0,127,255}
\definecolor{gttContextText}{RGB}{255,153,51}
\definecolor{gttDecoderText}{RGB}{0,153,0}

\newcommand{\gttenc}[1]{\textcolor{gttEncoderText}{#1}\xspace}
\newcommand{\gttctx}[1]{\textcolor{gttContextText}{#1}\xspace}
\newcommand{\gttdec}[1]{\textcolor{gttDecoderText}{#1}\xspace}

\newcommand{\gttphi}{\ensuremath{\textcolor{gttEncoderText}{\phi}}\xspace}
\newcommand{\gttS}{\ensuremath{\textcolor{gttContextText}{S}}\xspace}
\newcommand{\gttD}{\ensuremath{\textcolor{gttDecoderText}{D}}\xspace}

\title[Graph-to-Table Interfaces Control Relational Generalisation]
{Graph-to-Table Interfaces Control Relational Generalisation}

\author[A. Cagiano et al.]{%
Alessandro Cagiano \\
University of Milan \\
\email{alessandro.cagiano@unimi.it}
\And
Lorenzo Molfetta \\
University of Bologna \\
\email{lorenzo.molfetta@unibo.it}
\AND
Gianluca Moro \\
University of Bologna \\
\email{gianluca.moro@unibo.it}
\And
Matteo Zignani \\
University of Milan \\
\email{matteo.zignani@unimi.it}
\And
Pasquale Minervini \\
University of Edinburgh \\
Miniml.AI \\
\email{p.minervini@ed.ac.uk}
}

\begin{document}

\maketitle

\begin{abstract}
Graph-to-table systems combine three design choices: an encoder that maps relational evidence to feature rows, a context constructor that selects labelled examples, and a tabular decoder.
We investigate whether these components can be evaluated independently through matched comparisons that change one component while holding the rest of the pipeline and the evaluation constant.
Across logistic regression, \CatBoost, and \TabPFNthree on CLUTRR, 57 GraphLog worlds, and 57 knowledge graphs, we show that conclusions about one component can change when the others change.
On CLUTRR, absolute positional encodings maintain all distinctions among the evaluated queries and fit the short-path setting nearly perfectly; yet, they generalise worse than relation-count features because the supported prefix acts as a shortcut that breaks down on longer paths.
On GraphLog, encoder preference depends on the decoder, while the relation-count representation degrades as CatBoost tree depth increases.
Across different knowledge graphs, adding the frozen \Ultra score to local features is consistently useful, whereas adding local features to the score has a smaller and less stable effect.
A pointwise \TabPFNthree readout using the full encoder underperforms native \Ultra on 48 of 57 datasets.
These results show that encoder assessment depends jointly on context and decoder, while panel-level conclusions can also depend on evaluation and aggregation protocols.
\end{abstract}

\section{Introduction}

Prior-data fitted networks (PFNs) make predictions from a small labelled table without target-specific fine-tuning~\citep{hollmann2025tabpfn,grinsztajn2026tabpfn3}.
Their use is becoming increasingly common in graph-to-table approaches to relational prediction.
In this setting, a \gttenc{graph-to-table encoder} \gttphi maps each relational query and its surrounding graph evidence to a feature row, a \gttctx{context constructor} \gttS selects labelled rows from the target graph to form the prediction context, while a \gttdec{tabular decoder} \gttD infers the labels of held-out queries. We refer to the combination $(\gttphi,\gttS,\gttD)$ as the graph-to-table interface.

Recent work explores related approaches through graph-derived feature representations, graph-aware architectures, and learnt alignment between graph and tabular representations~\citep{eremeev2025g2tfm,eremeev2025graphpfn,liao2026tfmlinker,gao2026kgpfn,hu2026gtalign}. 
When such solutions are evaluated only through final predictive performance, it is difficult to determine where gains and failures originate.
The three components of this interface impose distinct constraints.
The \gttenc{encoder} determines which relational distinctions are preserved in each query row.
The \gttctx{context} determines which of these distinctions are associated with the target labels.
The \gttdec{decoder} determines which dependencies between the resulting features and labels can be exploited.
Consequently, preserving more detail does not by itself guarantee better generalisation.
Patterns predictive in the labelled context may not generalise to held-out queries, and the decoder may change the relative ranking of encoders.
End-to-end performance alone cannot distinguish among these effects.

We study how the \gttenc{encoder}, \gttctx{context}, and \gttdec{decoder} jointly determine relational generalisation.
We ask:
\begin{quote}
\emph{When does a graph-to-table representation generalise beyond its labelled context?}
\end{quote}
We investigate this question on controlled composition tasks and on a 57-knowledge-graph panel spanning multiple decoders, context constructions, and graph-derived encoders.\footnote{Code and artefacts are available at
\url{https://cagianoaless.github.io/tab_repres_gen/}.}

Our findings are as follows:
\begin{itemize}
\item We use paired diagnostics to determine whether an encoder preserves target-relevant features, whether those features transfer from the labelled support examples to held-out queries, and whether the decoder can utilise them.
\item We show on CLUTRR that absolute positions exploit a prefix shortcut that fails on longer paths, and on GraphLog that encoder preference changes across decoders and is sensitive to tree capacity.
\item On 57 knowledge graphs, we find that adding the frozen \Ultra{} score to local features is consistently useful, whereas adding local features to the score has a smaller and less stable effect.
Native-ranking results depend on panel weighting, while encoder effects depend on context construction.
\end{itemize}

\section{Related Work}

\paragraph{Graph--table foundation models.}
Building on tabular in-context learning with \TabPFN and \TabPFNthree~\citep{hollmann2025tabpfn,grinsztajn2026tabpfn3}, several methods expose graph structure to tabular predictors.
TabGFM~\citep{hayler2025graphs} augments node attributes with graph-derived structural features, whereas G2T-FM~\citep{eremeev2025g2tfm} constructs tabular representations through neighbourhood aggregation and structural encodings.
GTAlign~\citep{hu2026gtalign} instead learns an explicit alignment between graph and tabular representations.
These approaches establish interfaces between graph and tabular representations, while our focus is to isolate which query-level relational information must be preserved for link-level transfer.

\paragraph{In-context link prediction.}
In-context link prediction adapts a shared predictor to a target graph through labelled demonstrations, keeping the predictor's parameters fixed.
UniLP~\citep{dong2025unilp} reasons directly over graph neighbourhoods, TFMLinker~\citep{liao2026tfmlinker} represents candidate links as topology-aware rows with local and global contexts, and KGPFN~\citep{gao2026kgpfn} combines transferable relation representations with query-local features and global context.
These methods differ in both the underlying predictor and where relational evidence is introduced: through graph encoding, row construction, or demonstration selection.
Aggregate performance therefore leaves the contribution of each component unresolved. We address this distinction by varying path encoding and labelled context independently under a fixed evaluation protocol.

\paragraph{Length generalisation and contextual shortcuts.}
CLUTRR~\citep{sinha2019clutrr} evaluates relational composition over held-out kinship rules and chain lengths, whereas GraphLog~\citep{sinha2020graphlog} evaluates logical generalisation across graph worlds governed by different relational rules.
Such settings instantiate compositional generalisation~\citep{hupkes2020compositionality}, requiring learned components to remain productive under unseen combinations.
For Transformer models, length extrapolation is likewise sensitive to positional representation~\citep{kazemnejad2023position}.
These sensitivities relate to shortcut learning, where predictive regularities learned in one regime fail under distributional shift~\citep{geirhos2020shortcut}.
Within tabular in-context learning, VIP-COP~\citep{chen2026vipcop} studies the selection of informative rows and features under a limited context budget, while OpenRFM~\citep{chen2026openrfm} identifies relation-level label coverage as a critical factor for relational inference.
We examine these sources of failure jointly by evaluating across path lengths while independently controlling the encoder and labelled context.

\paragraph{Systematic component analyses.}
Several studies show that model comparisons can change substantially when architecture, training, and evaluation choices are controlled separately.
In knowledge graph completion, \citet{ruffinelli2020old} systematically study the effects of model architecture, training strategy, and hyperparameter optimisation, while changes in the training regime can substantially alter comparisons among knowledge graph embedding models~\citep{jain2020baseline}.
In graph learning, GraphGym~\citep{you2020designspace} studies a modular graph neural network (GNN) design space under a unified evaluation setup, separating architectural choices that are often compared only through complete models.
Our work follows this controlled-comparison perspective, but applies it to graph-to-table prediction by varying the encoder, labelled context, and decoder while keeping the remaining pipeline and evaluation protocol fixed.

\paragraph{Representation expressivity and distinguishability.}
Relational expressivity has been studied through supported relational patterns and structural motifs~\citep{trouillon2019inductive,huang2025motif}, while Weisfeiler--Leman analyses characterise graph representations through structures they cannot distinguish~\citep{morris2019weisfeiler,xu2019powerful}.
Our row-equivalence diagnostic applies the same principle after graph-to-table conversion.
If two labelled queries induce identical rows within the same prediction context, they are necessarily indistinguishable to the tabular decoder.
The converse does not hold: distinguishable encodings merely remove this representational obstruction, since successful transfer additionally depends on preserving task-relevant information and maintaining adequate source--target support~\citep{johansson2019support}.

\paragraph{Inductive and zero-shot knowledge graph reasoning.}
Inductive knowledge graph methods use transferable structural computations, including enclosing subgraphs in GraIL~\citep{teru2020grail}, path-based reasoning in NBFNet~\citep{zhu2021nbfnet}, and relation graphs in InGram~\citep{lee2023ingram}.
\Ultra{}~\citep{galkin2024ultra} extends structural reasoning to zero-shot transfer across knowledge graphs with previously unseen relation vocabularies, while TRIX~\citep{zhang2025trix} develops more expressive fully inductive triplet representations for entity and relation prediction.
Our setting keeps \Ultra frozen and exposes its candidate score as a tabular feature, testing whether local graph evidence adds information beyond the native score and whether in-context tabular decoding improves the resulting ranking.
\begin{figure*}[t]
\centering
\includegraphics[width=\textwidth]{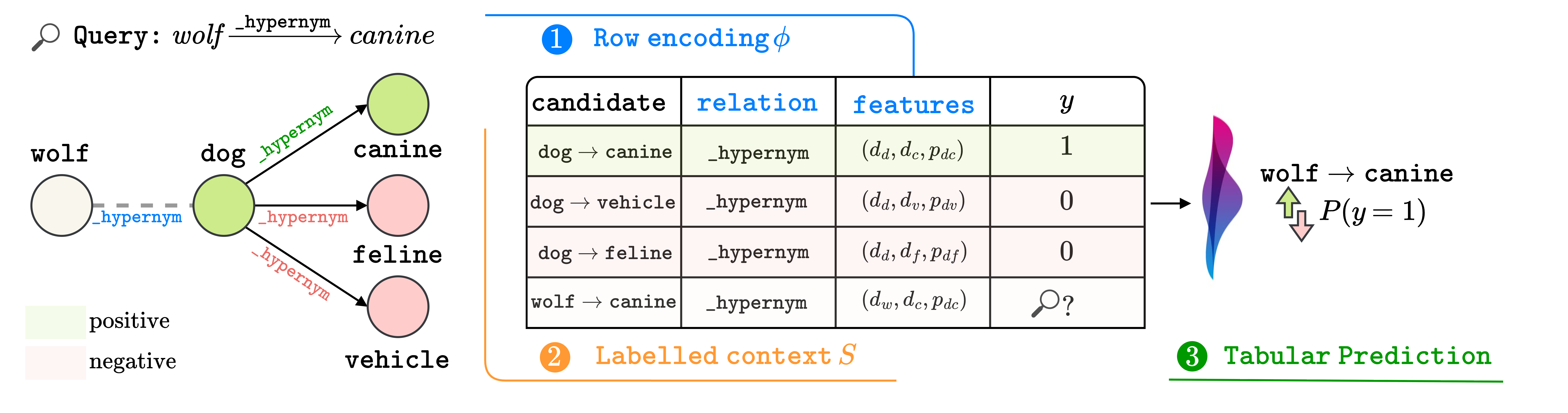}
\caption{\textbf{Graph-to-table prediction pipeline.}
Given a relational query and its surrounding graph, the encoder $\gttphi$
constructs a feature row, the context constructor $\gttS$ assembles labelled examples
from the target graph, and the tabular decoder $\gttD$ produces prediction scores
for the held-out query.}
\label{fig:audit}
\end{figure*}

\section{Graph-to-Table Diagnostic Framework} \label{sec:audit}
A graph-to-table pipeline comprises three components: an encoder \(\gttphi\) that maps graph evidence to a feature row, a context constructor \(\gttS\) that assembles labelled rows, and a decoder \(\gttD\) that predicts from the resulting context (see \cref{fig:audit}).
End-to-end performance conflates the effects of these components.
We therefore organise the analysis around four diagnostic questions concerning row distinguishability, feature alignment, decoder dependence, and query-level ranking effects.
Each question either establishes an encoder-induced ceiling or compares matched conditions while holding the remaining pipeline and evaluation protocol fixed.

\subsection{Problem Formulation}
Let \(G=(V,R,E)\) be an observed relational graph, where \(V\) is the set of entities, \(R\) is the set of relation types, and \(E\subseteq V\times R\times V\) is the set of observed edges.
A candidate query \(q=(h,r,t)\) associates a head entity \(h\in V\) and a tail entity \(t\in V\) through relation \(r\in R\).
Let \(\mathcal{G}\), \(\mathcal{Q}\), \(\mathcal{X}\), and \(\mathcal{Y}\) denote the graph, query, feature-row, and label spaces, respectively.
The encoder
\[
\gttphi:\mathcal{G}\times\mathcal{Q}\rightarrow\mathcal{X}
\]
maps a graph and query to a feature row \(x=\gttphi(G,q)\).
Given a pool of labelled rows and a requested budget \(n\in\mathbb{N}\), i.e., the number of labelled rows to include per class, the context constructor
\[
\gttS:(\mathcal{X}\times\mathcal{Y})^{*}\times\mathbb{N}\rightarrow\mathcal{C}
\]
assembles a labelled context \(\context_n\), where \(\mathcal{C}\) denotes the space of possible contexts.
Finally, the decoder
\[
\gttD:\mathcal{C}\times\mathcal{X}\rightarrow\mathbb{R}^{|\mathcal{Y}|}
\]
maps the labelled context and a query row to class scores.
We obtain a predicted label from the highest-scoring class, while in binary link prediction we can use the positive-class score for candidate ranking.
For example, in our path-based experiments, \(\gttphi\) may encode a relational path through relation counts or absolute positions, \(\gttS\) selects labelled rows from the target graph, and \(\gttD\) is instantiated by logistic regression, CatBoost, or TabPFN-3.
The resulting pipeline can be written as
\begin{equation}
\context_n
=
\gttS\!\left(
    \{(\gttphi(G,q_i),y_i)\}_{i\in\mathcal I},
    n
\right),
\qquad
\mathbf{s}
=
\gttD(\context_n,x),
\qquad
\hat y
=
\arg\max_{y\in\mathcal{Y}} s_y.
\label{eq:pipeline}
\end{equation}

\subsection{Diagnostic Questions}

\paragraph{\texorpdfstring{\protect\faClone\hspace{0.35em}}{}%
Does the row distinguish queries with different labels?}
Before considering any particular decoder, we ask whether the encoder maps queries with different labels to distinguishable rows.
For a fixed graph \(G\), the encoder partitions queries according to row identity: we write
\(q\sim_{\gttphi}q'\) if and only if
\(\gttphi(G,q)=\gttphi(G,q')\).
Queries in the same equivalence class expose the same row to the decoder and, with the context and remaining inputs fixed, cannot be distinguished from that row alone.
To quantify this representational limitation independently of the decoder, we compute an oracle accuracy ceiling on the finite set of evaluated queries, each paired with its label.
Let \(N\) be the number of such queries, \(n_{xy}\) the number of queries mapped to row \(x\) with label \(y\), and \(n_x=\sum_y n_{xy}\) the number of queries mapped to row \(x\).
Any predictor that observes only \(x\) must make the same prediction for all examples sharing that row, and can therefore classify at most \(\max_y n_{xy}\) of them correctly.
The resulting empirical ceiling is
\begin{equation} \label{eq:ceiling}
\Astar(\gttphi)
=
\frac{1}{N}\sum_x\max_y n_{xy}
=
\sum_x p(x)\max_y p(y\mid x),
\end{equation}
\noindent where \(p(x)\) and \(p(y\mid x)\) are empirical frequencies.
Under zero--one loss, this is the empirical Bayes-optimal accuracy obtained by predicting the majority label within each row class~\citep{devroye1996pattern}.
Evaluation labels are used only to quantify ambiguity introduced by the encoder.
Successful transfer additionally requires the decoder to infer an appropriate rule from the available labelled context.

A ceiling of one does not necessarily imply that the representation resolves repeated query patterns: it may also arise because every evaluated query maps to a unique row.
We therefore report the singleton mass
\begin{equation} \label{eq:singleton-mass}
\sigma_{\gttphi}
=
\sum_x p(x)\mathbf{1}[n_x=1],
\end{equation}
where \(n_x\) is the number of queries mapped to row \(x\).
A value of \(\sigma_{\gttphi}=1\) indicates that every evaluated query occupies a singleton equivalence class.
Since singleton rows are necessarily assigned a ceiling accuracy of one in \cref{eq:ceiling}, they can mask ambiguity among rows that occur more than once.
We therefore condition the ceiling on repeated rows:
\begin{equation}
A_{\mathrm{ns}}^\star(\gttphi)
=
\frac{\sum_{x:n_x>1}p(x)\max_y p(y\mid x)}
     {1-\sigma_{\gttphi}},
\qquad
A^\star(\gttphi)
=
\sigma_{\gttphi}
+
(1-\sigma_{\gttphi})A_{\mathrm{ns}}^\star(\gttphi).
\label{eq:nonsingleton-ceiling}
\end{equation}
The conditional quantity is undefined when \(\sigma_{\gttphi}=1\); it measures the maximum row-only accuracy among examples belonging to repeated equivalence classes, separating ambiguity within repeated rows from the trivial contribution of singleton rows.

For GraphLog, evaluation uses macro balanced accuracy, which weights every class equally.
For a row \(x\), predicting class \(y\) contributes \(n_{xy}/n_y\) to the class-normalised number of correct predictions.
The optimal row-only prediction therefore chooses the class maximising this quantity, yielding the corresponding class-reweighed ceiling:
\begin{equation} \label{eq:balanced-ceiling}
B^\star(\gttphi)
=
\frac{1}{K}
\sum_x
\max_y
\frac{n_{xy}}{n_y},
\end{equation}
where \(K\) is the number of observed classes, \(n_{xy}\) counts examples with row \(x\) and label \(y\), and \(n_y\) is the number of examples in class \(y\).

\paragraph{\texorpdfstring{\protect\faLink\hspace{0.35em}}{}%
Does feature alignment improve predictions on unseen queries?}
Let \(b(q)\) denote an additional block of related feature columns associated with query \(q\).
We compare three matched conditions: (i) the \emph{base} encoder excludes the block, (ii) the \emph{permuted} encoder includes the same collection of complete blocks but permutes their correspondence to queries, and (iii) the \emph{aligned} encoder restores the original query-to-block correspondence.
The \emph{base}--\emph{permuted} comparison measures the effect of adding the block without preserving its query alignment.
The \emph{permuted}--\emph{aligned} comparison measures the effect of restoring that alignment while holding fixed the block dimensionality, the collection of block values, context construction, and decoder.
This comparison separates sensitivity to query--feature alignment from the mere presence of the additional feature block.

\paragraph{\texorpdfstring{\protect\faCogs\hspace{0.35em}}{}%
Does encoder preference change across decoders?}
For two encoders \(\gttphi_1\) and \(\gttphi_2\), we compare their performance
under the same context construction and evaluation queries.
For a decoder \(\gttD\), let
\[
\Delta_{\gttD}
=
M(\gttphi_1,\gttS,\gttD)
-
M(\gttphi_2,\gttS,\gttD),
\]
where \(M\) denotes the task metric: a change in the sign of \(\Delta_{\gttD}\) across decoders indicates a reversal
in the relative ranking of the two encoders.
We also inspect each encoder's absolute performance, since a reversal can arise from degradation in one representation or improvement in the other.

\paragraph{\texorpdfstring{\protect\faSortNumericDown\hspace{0.35em}}{}%
For which query regimes does graph evidence improve ranking?}
Aggregate ranking metrics may conceal heterogeneous effects across query types.
We therefore partition evaluation queries using graph evidence observable before prediction, such as the presence of direct or reverse edges between the query endpoints.
Within each stratum, we compare reciprocal rank across encoder variants on the same queries.
These comparisons keep the context construction, decoder, candidate set, filtering protocol, and random seed fixed.
The resulting stratification identifies the query regimes in which particular graph-derived feature groups are associated with ranking improvements.
\section{Experimental Design}
\label{sec:experiments}
We evaluate the diagnostic framework in three regimes: synthetic calibration, controlled compositional extrapolation, and heterogeneous cross-graph ranking.
Whenever the decoder \(\gttD\) is varied, logistic regression, \CatBoost{}, and frozen \TabPFNthree{} receive the same queries, labels, context examples, and feature values before decoder-specific categorical encoding.
Decoder configurations remain fixed within each matched comparison.

\subsection{Calibration and Compositional Evaluation}

\paragraph{Controlled calibration.}
We generate four binary tasks governed by reciprocity, typed relation composition, local structural roles, and shared graph-community membership.
For each task, ten independently generated graphs define disjoint context and evaluation populations.
Because the relevant relational distinctions are known by construction, these tasks validate the row-equivalence ceiling, singleton mass, and matched-permutation diagnostics before applying them to benchmark data.
\cref{app:controlled} provides the complete generation procedure and results.

In CLUTRR, GraphLog, and the dihedral control, each row represents a supplied or generated symbolic sequence.
The encoders differ only in which properties of that sequence are exposed to the decoder.

\paragraph{CLUTRR}
We use the official \textit{gen-train23-test2to10} split~\citep{sinha2019clutrr}.
Contexts contain training paths of lengths 2--3 with nested budgets of \(5,10,20,40,80\) examples per target relation, while evaluation covers lengths 4--10.
We compare three encoders of the same symbolic path: a \emph{relation-multiset} encoder, recording path length and relation counts; an \emph{absolute-position} encoder, preserving the relation at each position; and a \emph{directed-transition} encoder, counting ordered adjacent relation pairs without absolute positions.
For the alignment test, we append the absolute-position block to the relation-multiset row and compare the correctly aligned block with a condition in which complete blocks are permuted among paths of equal length.
This isolates the effect of query--feature alignment while preserving the block dimensionality and collection of values.
A separate suffix ablation zeros absolute-position coordinates beyond position~3, testing whether unsupported positions contribute to the observed behaviour.
To isolate length coverage from context volume, the same 80-example length-2--3 base context receives either 20 further short examples or 20 length-4 examples, yielding 100 examples per target relation in both conditions.
\cref{app:clutrr} reports the schemas, dimensionalities, and split statistics.

\paragraph{GraphLog.}
We use the official symbolic path descriptors from all 57 worlds~\citep{sinha2020graphlog}.
Contexts contain paths of lengths 2--3 with nested budgets of \(5,10,20,40\) examples per target class, while evaluation uses all test paths of lengths 4--5.
All worlds share the same vocabulary of 20 directed relation symbols but differ in their relational rules.
We compare \emph{relation-multiset} and \emph{absolute-position} encoders.
Neither includes entity identifiers or the target relation, so the comparison isolates how the supplied symbolic path is represented.
Performance is measured using macro balanced accuracy together with the corresponding class-balanced row-equivalence ceiling.
\cref{app:composition} gives the complete schemas, alignment controls, and decoder sensitivities.

\paragraph{Dihedral boundary control.}
The dihedral control contains ten independently generated populations over the 16 symmetries of a regular octagon.
Context sequences have lengths 2--3, evaluation sequences have lengths 4--10, and the target is their ordered group product.
We compare an \emph{absolute-position} encoder with a \emph{coarse} encoder that removes sequence order and merges group elements separated by four rotations.
This provides a boundary case in which coarsening discards task-relevant information, testing whether the preference observed for a coarser representation on CLUTRR persists when the removed information is required by the target.

\subsection{Knowledge Graph Panel}

\paragraph{Graphs and ranking protocol.}
The panel contains all \UltraDatasets{} knowledge graphs, grouped into \UltraFamilies{} provenance families, with \Ultra{}~\citep{galkin2024ultra} frozen throughout.
For each graph, we sample 64 official test triples stratified by relation and evaluate both head and tail prediction by ranking all entities under standard filtered evaluation.
For every candidate entity, the resulting candidate triple is mapped by the encoder to one feature row.
All graph-derived features are computed from the observed support graph, which excludes the held-out target triples, and context positives are disjoint from the sampled evaluation queries.
This keeps the candidate-ranking problem fixed while varying the graph evidence exposed to the tabular decoder.

\paragraph{Encoder variants.}
We evaluate the nine encoder variants summarised in Appendix \cref{tab:interfaces}.
They include \emph{Entity tokens}, exposing dataset-scoped endpoint tokens; \emph{Dyadic}, based on evidence directly connecting the query endpoints; and \emph{Paths}, based on two-hop path evidence.
The broader \emph{Local} encoder combines relation identity with endpoint, pair, path, and community statistics, whereas \emph{Global} exposes graph-wide topological and spectral information.
\emph{ULTRA score} exposes only the candidate score produced by the frozen graph-native \Ultra{} model.
Composite encoders concatenate these fixed blocks as \emph{Local + Global}, \emph{Local + ULTRA score}, and \emph{Full}.
The complete encoder ladder is evaluated with nested context budgets of \(20,50,100,200\) examples per binary class.
Its baseline negatives are endpoint corruptions without an all-known-positive rejection step. \cref{app:ultra57} separately isolates the effect of this filtering choice with a matched control.
Whenever the frozen ULTRA score is included, we also report the native ULTRA candidate ranking as a reference.

\paragraph{Pointwise-ranking control.}
The tabular decoders are trained with a pointwise binary objective but are evaluated by filtered ranking within each directional query.
At 200 examples per class, we therefore cross the \emph{Local}, \emph{ULTRA score}, and \emph{Local + ULTRA score} encoders with raw, within-query z-score, and within-query percentile representations of their numerical columns.
The transformations are label-free and applied by the same rule to context groups and evaluation candidate blocks.
This tests whether query-relative normalisation changes the behaviour of the pointwise readouts under an otherwise fixed ranking protocol.
\cref{app:query-relative} gives the grouping and results.

\paragraph{Negative-context construction.}
For the negative-context comparison, we fix the budget at 200 examples per class and vary only the context constructor \(\gttS\), while keeping context positives and evaluation queries fixed.
\emph{Filtered-uniform} negatives replace the missing endpoint with a uniformly sampled entity outside the known-positive set.
\emph{Structurally plausible} negatives favour entities observed in the appropriate relation role and, when possible, in the same graph partition as the fixed endpoint, followed by degree similarity.
Selection uses only graph-based criteria, so the comparison changes the difficulty and structure of the labelled negative context without using evaluation labels or the frozen ULTRA score.
\cref{app:ultra57} gives the complete construction and audit checks.

\paragraph{Positive-context construction.}
We additionally compare \emph{pool-random positives} with \emph{evaluation-relation-matched positives} while keeping context size and evaluation queries fixed.
The latter selects positives to better match the relation distribution represented by the evaluation queries, allowing us to vary relation-level support without changing the evaluation set.

\subsection{Statistical Analysis}

For the 57-KG panel, the metric is averaged over queries and seeds within datasets, then over datasets within provenance families, with families weighted equally.
We refer to this estimand as the \emph{equal-family} metric and additionally report equal-dataset aggregation as a sensitivity analysis.
Provenance families are also the unit of inference: bootstrap sensitivity intervals resample families with replacement, and the exact max-\(|t|\) multiple-comparison correction in \cref{app:paired-context-effects} flips the sign of each family's mean paired effect as a block.
Because the panel is fixed rather than sampled from a population of graphs, we read bootstrap intervals as sensitivity to panel composition rather than as confidence intervals, and leave-one-family-out ranges show how much each estimate depends on any single family.
For CLUTRR, effects are paired across context seeds on the same official split. GraphLog comparisons are paired across worlds under shared matched conditions. The dihedral control uses the independently generated populations as replication units.
Query-level ranking analyses stratify the same evaluation queries by graph evidence observable before prediction and report reciprocal rank within each stratum.
\section{Results}
\label{sec:results}

\subsection{Aligned Prefixes Form a Shortcut Under Length Shift}
\label{sec:clutrr}

At 80 context examples per target relation, the absolute-position encoder reaches \ClutrrLogAbsolute{}/\ClutrrCatAbsolute{}/\ClutrrTabAbsolute{} accuracy for logistic regression, \CatBoost{}, and \TabPFNthree{}, whereas the relation-multiset encoder reaches \ClutrrLogMultiset{}/\ClutrrCatMultiset{}/\ClutrrTabMultiset{}.
The absolute-position ceiling of \ClutrrAbsoluteCeiling{} follows from its singleton mass of \ClutrrAbsoluteSingleton{} and therefore carries no information about ambiguity among repeated rows.
For the relation multiset, the raw ceiling is \ClutrrMultisetCeiling{}, while the repeated-row mass is \ClutrrMultisetNonSingletonMass{} and its conditional ceiling is \ClutrrMultisetNonSingletonCeiling{}.
The lower absolute-position accuracy therefore occurs despite injective rows, while its perfect ceiling follows from that injectivity.
The multiset advantage also holds across all declared metrics and for at least 15 of the 18 target relations (\cref{app:clutrr}, \crefrange{tab:clutrr_levels}{tab:clutrr_metric_robustness}).

\begin{figure*}[t]
\centering
\resizebox{0.94\textwidth}{!}{\begingroup
\fontfamily{ptm}\selectfont
\definecolor{plotbackground}{HTML}{FAF5F3}%
\definecolor{gray}{HTML}{394144}%
\definecolor{blue-1}{HTML}{4D4DFF}%
\definecolor{blue-2}{HTML}{8F8FFF}%
\definecolor{orange-1}{HTML}{FBA418}%
\definecolor{orange-2}{HTML}{FDCC7B}%
\definecolor{green-1}{HTML}{F00000}%
\definecolor{green-2}{HTML}{FF7575}%
\definecolor{red-1}{HTML}{F00000}%
\definecolor{red-2}{HTML}{FF7575}%
\pgfdeclarehorizontalshading{bluegradient}{100bp}{%
  color(0bp)=(blue-1);%
  color(30bp)=(blue-1);%
  color(60bp)=(blue-2);%
  color(100bp)=(blue-2)%
}%
\pgfdeclarehorizontalshading{orangegradient}{100bp}{%
  color(0bp)=(orange-1);%
  color(30bp)=(orange-1);%
  color(60bp)=(orange-2);%
  color(100bp)=(orange-2)%
}%
\pgfdeclarehorizontalshading{greengradient}{100bp}{%
  color(0bp)=(green-1);%
  color(30bp)=(green-1);%
  color(60bp)=(green-2);%
  color(100bp)=(green-2)%
}%
\pgfdeclareverticalshading{redgradient}{100bp}{%
  color(0bp)=(red-1);%
  color(30bp)=(red-1);%
  color(60bp)=(red-2);%
  color(100bp)=(red-2)%
}\colorlet{decoder0}{blue!80!magenta!60!white}%
\definecolor{decoder1}{HTML}{E0CB68}%
\colorlet{decoder2}{red!75!orange}%
\begin{tikzpicture}
\begin{groupplot}[group style={group size=3 by 1,horizontal sep=19mm},
width=4.2cm,height=4.2cm,y dir=reverse,xmajorgrids=true,ymajorgrids=false,
grid=both,grid style=dashed,axis background/.style={fill=plotbackground},
    every tick label/.append style={font=\fontsize{8}{8}\selectfont},
    scaled ticks=false,tick label style={/pgf/number format/fixed,/pgf/number format/precision=2},
    xlabel style={font=\small},ylabel style={font=\small},title style={font=\small,yshift=2pt},label style={font=\fontsize{8}{8}\selectfont}]
\nextgroupplot[clip=false,title={(a) OOD accuracy},xmin=0,xmax=.6,ymin=-.5,ymax=2.5,xtick={0,.2,.4,.6},ytick={0,1,2},yticklabels={{Multiset},{+ permuted},{+ aligned}},xlabel={Accuracy}]
\addplot[xbar,bar width=5pt,bar shift=0pt,fill=decoder0,draw=black,line width=.35pt] coordinates {(0.4295186274,-0.2300000000) (0.3996985490,0.7700000000) (0.2663183266,1.7700000000)};
\addplot[xbar,bar width=5pt,bar shift=0pt,fill=decoder1,draw=black,line width=.35pt] coordinates {(0.2725309267,0.0000000000) (0.2554834756,1.0000000000) (0.1629944184,2.0000000000)};
\addplot[xbar,bar width=5pt,bar shift=0pt,fill=decoder2,draw=black,line width=.35pt] coordinates {(0.2893819084,0.2300000000) (0.2917053467,1.2300000000) (0.1874272680,2.2300000000)};
\node[anchor=west,inner sep=0pt,font=\fontsize{5}{5}\selectfont,text=black] at ([xshift=2pt,yshift=0pt]axis cs:0.4295186274,-0.2300000000) {\pgfmathprintnumber[fixed,fixed zerofill,precision=3]{0.4295186274}};
\node[anchor=west,inner sep=0pt,font=\fontsize{5}{5}\selectfont,text=black] at ([xshift=2pt,yshift=0pt]axis cs:0.3996985490,0.7700000000) {\pgfmathprintnumber[fixed,fixed zerofill,precision=3]{0.3996985490}};
\node[anchor=west,inner sep=0pt,font=\fontsize{5}{5}\selectfont,text=black] at ([xshift=2pt,yshift=0pt]axis cs:0.2663183266,1.7700000000) {\pgfmathprintnumber[fixed,fixed zerofill,precision=3]{0.2663183266}};
\node[anchor=west,inner sep=0pt,font=\fontsize{5}{5}\selectfont,text=black] at ([xshift=2pt,yshift=0pt]axis cs:0.2725309267,0.0000000000) {\pgfmathprintnumber[fixed,fixed zerofill,precision=3]{0.2725309267}};
\node[anchor=west,inner sep=0pt,font=\fontsize{5}{5}\selectfont,text=black] at ([xshift=2pt,yshift=0pt]axis cs:0.2554834756,1.0000000000) {\pgfmathprintnumber[fixed,fixed zerofill,precision=3]{0.2554834756}};
\node[anchor=west,inner sep=0pt,font=\fontsize{5}{5}\selectfont,text=black] at ([xshift=2pt,yshift=0pt]axis cs:0.1629944184,2.0000000000) {\pgfmathprintnumber[fixed,fixed zerofill,precision=3]{0.1629944184}};
\node[anchor=west,inner sep=0pt,font=\fontsize{5}{5}\selectfont,text=black] at ([xshift=2pt,yshift=0pt]axis cs:0.2893819084,0.2300000000) {\pgfmathprintnumber[fixed,fixed zerofill,precision=3]{0.2893819084}};
\node[anchor=west,inner sep=0pt,font=\fontsize{5}{5}\selectfont,text=black] at ([xshift=2pt,yshift=0pt]axis cs:0.2917053467,1.2300000000) {\pgfmathprintnumber[fixed,fixed zerofill,precision=3]{0.2917053467}};
\node[anchor=west,inner sep=0pt,font=\fontsize{5}{5}\selectfont,text=black] at ([xshift=2pt,yshift=0pt]axis cs:0.1874272680,2.2300000000) {\pgfmathprintnumber[fixed,fixed zerofill,precision=3]{0.1874272680}};
\nextgroupplot[clip=false,title={(b) Paired changes},xmin=-.25,xmax=.35,ymin=-.5,ymax=1.5,xtick={-.1,0,.1,.2},ytick={0,1},yticklabels={{Block addition},{Alignment cost}},xlabel={Accuracy change}]
\addplot[mark=none,bar shift=0pt,xbar,bar width=6pt,fill=decoder0,draw=black,line width=.35pt] coordinates {(-0.0298200784,-0.2200000000) (0.1333802224,0.7800000000)};
\addplot[mark=none,bar shift=0pt,xbar,bar width=6pt,fill=decoder1,draw=black,line width=.35pt] coordinates {(-0.0170474511,0.0000000000) (0.0924890572,1.0000000000)};
\addplot[mark=none,bar shift=0pt,xbar,bar width=6pt,fill=decoder2,draw=black,line width=.35pt] coordinates {(0.0023234383,0.2200000000) (0.1042780787,1.2200000000)};
\draw[black!60,line width=.5pt] (axis cs:0,-0.5) -- (axis cs:0,1.5);
\node[anchor=east,inner sep=0pt,font=\fontsize{5}{5}\selectfont,text=black] at ([xshift=-2pt,yshift=0pt]axis cs:-0.0298200784,-0.2200000000) {\pgfmathprintnumber[fixed,fixed zerofill,precision=3]{-0.0298200784}};
\node[anchor=west,inner sep=0pt,font=\fontsize{5}{5}\selectfont,text=black] at ([xshift=2pt,yshift=0pt]axis cs:0.1333802224,0.7800000000) {\pgfmathprintnumber[fixed,fixed zerofill,precision=3]{0.1333802224}};
\node[anchor=east,inner sep=0pt,font=\fontsize{5}{5}\selectfont,text=black] at ([xshift=-2pt,yshift=0pt]axis cs:-0.0170474511,0.0000000000) {\pgfmathprintnumber[fixed,fixed zerofill,precision=3]{-0.0170474511}};
\node[anchor=west,inner sep=0pt,font=\fontsize{5}{5}\selectfont,text=black] at ([xshift=2pt,yshift=0pt]axis cs:0.0924890572,1.0000000000) {\pgfmathprintnumber[fixed,fixed zerofill,precision=3]{0.0924890572}};
\node[anchor=west,inner sep=0pt,font=\fontsize{5}{5}\selectfont,text=black] at ([xshift=2pt,yshift=0pt]axis cs:0.0023234383,0.2200000000) {\pgfmathprintnumber[fixed,fixed zerofill,precision=3]{0.0023234383}};
\node[anchor=west,inner sep=0pt,font=\fontsize{5}{5}\selectfont,text=black] at ([xshift=2pt,yshift=0pt]axis cs:0.1042780787,1.2200000000) {\pgfmathprintnumber[fixed,fixed zerofill,precision=3]{0.1042780787}};
\nextgroupplot[clip=false,title={(c) Length exposure},xmin=-.8,xmax=1.5,ymin=-.5,ymax=1.5,xtick={0,.5,1},ytick={0,1},yticklabels={{Length 4},{Lengths 5--10}},xlabel={Accuracy change}]
\addplot[mark=none,bar shift=0pt,xbar,bar width=6pt,fill=decoder0,draw=black,line width=.35pt] coordinates {(0.4891774892,-0.2200000000) (-0.0006534514,0.7800000000)};
\addplot[mark=none,bar shift=0pt,xbar,bar width=6pt,fill=decoder1,draw=black,line width=.35pt] coordinates {(0.3722943723,0.0000000000) (0.0099134902,1.0000000000)};
\addplot[mark=none,bar shift=0pt,xbar,bar width=6pt,fill=decoder2,draw=black,line width=.35pt] coordinates {(0.7575757576,0.2200000000) (0.0326492675,1.2200000000)};
\draw[black!60,line width=.5pt] (axis cs:0,-0.5) -- (axis cs:0,1.5);
\node[anchor=west,inner sep=0pt,font=\fontsize{5}{5}\selectfont,text=black] at ([xshift=2pt,yshift=0pt]axis cs:0.4891774892,-0.2200000000) {\pgfmathprintnumber[fixed,fixed zerofill,precision=3]{0.4891774892}};
\node[anchor=east,inner sep=0pt,font=\fontsize{5}{5}\selectfont,text=black] at ([xshift=-2pt,yshift=0pt]axis cs:-0.0006534514,0.7800000000) {\pgfmathprintnumber[fixed,fixed zerofill,precision=3]{-0.0006534514}};
\node[anchor=west,inner sep=0pt,font=\fontsize{5}{5}\selectfont,text=black] at ([xshift=2pt,yshift=0pt]axis cs:0.3722943723,0.0000000000) {\pgfmathprintnumber[fixed,fixed zerofill,precision=3]{0.3722943723}};
\node[anchor=west,inner sep=0pt,font=\fontsize{5}{5}\selectfont,text=black] at ([xshift=2pt,yshift=0pt]axis cs:0.0099134902,1.0000000000) {\pgfmathprintnumber[fixed,fixed zerofill,precision=3]{0.0099134902}};
\node[anchor=west,inner sep=0pt,font=\fontsize{5}{5}\selectfont,text=black] at ([xshift=2pt,yshift=0pt]axis cs:0.7575757576,0.2200000000) {\pgfmathprintnumber[fixed,fixed zerofill,precision=3]{0.7575757576}};
\node[anchor=west,inner sep=0pt,font=\fontsize{5}{5}\selectfont,text=black] at ([xshift=2pt,yshift=0pt]axis cs:0.0326492675,1.2200000000) {\pgfmathprintnumber[fixed,fixed zerofill,precision=3]{0.0326492675}};
\end{groupplot}\node[anchor=north,font=\fontsize{8}{9}\selectfont,align=center] at ([yshift=-9.2mm]group c2r1.south) {\tikz[baseline=-.3ex]{\path[fill=decoder0,draw=black,line width=.3pt] (0,0) rectangle (1.2em,.65em);} Logistic\quad\tikz[baseline=-.3ex]{\path[fill=decoder1,draw=black,line width=.3pt] (0,0) rectangle (1.2em,.65em);} CatBoost\quad\tikz[baseline=-.3ex]{\path[fill=decoder2,draw=black,line width=.3pt] (0,0) rectangle (1.2em,.65em);} TabPFN-3};\end{tikzpicture}%
\endgroup}
\caption{\textbf{Effect of path-feature alignment and length exposure on CLUTRR.} (\textit{a}) Out-of-distribution (OOD) accuracy for the relation multiset alone, with a permuted absolute-position block, and with the correctly aligned block.
(\textit{b}) Paired accuracy changes from adding the permuted block and restoring its path-level alignment.
(\textit{c}) Accuracy change after replacing 20 additional length-2--3 examples with 20 length-4 examples, evaluated at length 4 and lengths 5--10.
Panels (\textit{a}) and (\textit{b}) use 80 context examples per target relation and evaluation paths of lengths 4--10, while panel (\textit{c}) uses the same 80-example base plus 20 additional examples, for 100 examples per target relation in both arms.
In (\textit{b}), positive block-addition values indicate higher OOD accuracy after adding the permuted block, while positive alignment values indicate accuracy lost after restoring the correct path-level alignment.
In (\textit{c}), positive values indicate higher OOD accuracy after adding length-4 context examples. Bar labels are rounded to three decimals.}
\label{fig:clutrr}
\end{figure*}
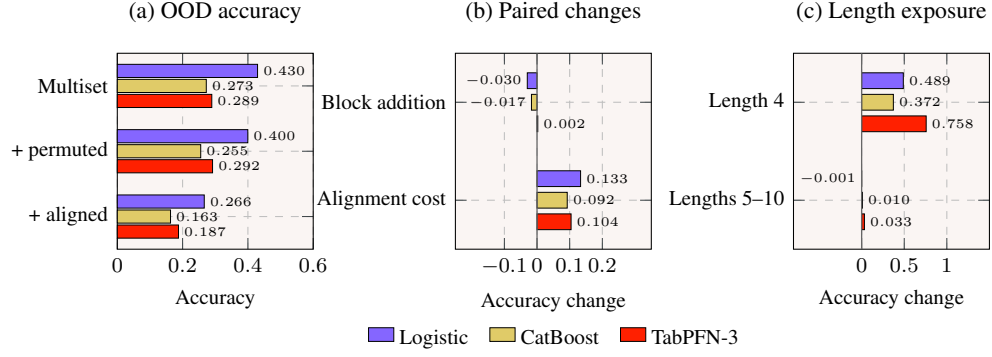

\cref{fig:clutrr} reports the matched alignment and length-exposure interventions under the short-to-long support shift.
Relative to the relation-multiset encoder, adding an absolute-position block permuted among paths of equal length changes accuracy by only \(-0.030/-0.017/+0.002\) for logistic regression, \CatBoost{}, and \TabPFNthree{}.
Restoring the correct path-level alignment reduces accuracy by a further \ClutrrLogAlignCost{}/\ClutrrCatAlignCost{}/\ClutrrTabAlignCost{}.
Context paths have lengths 2--3, so positions 4--10 have no labelled support. Evaluation paths have lengths 4--10.
Setting these unsupported suffix coordinates to zero changes predictions by less than \(3\times10^{-17}\).
The aligned encoders reach \ClutrrContextFitRange{} accuracy on the labelled context rows, while transferring substantially worse to longer paths.
These matched interventions therefore support a shortcut interpretation in which alignment exposes a highly predictive supported prefix that does not transfer to longer paths.

The length-exposure intervention produces a similarly local effect.
Replacing 20 additional short examples with 20 length-4 examples improves length~4 by \ClutrrAddLogFour{}/\ClutrrAddCatFour{}/\ClutrrAddTabFour{}, but changes lengths 5--10 by only \ClutrrAddLogLong{}/\ClutrrAddCatLong{}/\ClutrrAddTabLong{}.
The directed-transition encoder reaches \ClutrrLogTransition{}/\ClutrrCatTransition{}/\ClutrrTabTransition{}, with its ordering relative to the multiset changing across decoders.
A shared-parameter gated recurrent unit (GRU) with 32-dimensional relation embeddings, 64 hidden units, one recurrent layer, and context-only early stopping improves from \ClutrrGruContextLow{} to \ClutrrGruContextHigh{} OOD accuracy across the five context budgets.
At the largest budget it decreases from \ClutrrGruLengthFour{} at length~4 to \ClutrrGruLengthTen{} at length~10, with aggregate accuracy \ClutrrGru{}.
The GRU serves as an existence control for the split, showing that substantially stronger short-to-long transfer is attainable.
Together, these results support a shortcut interpretation for the tested tabular interfaces under the known length-support shift.

\subsection{A Decoder-Dependent Failure Regime}
\label{sec:decoder-dependence}

GraphLog reverses the encoder ordering observed on CLUTRR for the nonlinear decoders.
The performance difference between the relation multiset and absolute positions is \GraphLogLogDelta{} for logistic regression, \GraphLogCatDelta{} for \CatBoost{}, and \GraphLogTabDelta{} for \TabPFNthree{}.
Complete absolute levels are reported in \cref{tab:composition_levels}.
Both encoders retain high class-balanced ceilings of \GraphLogSharedCeiling{} and \GraphLogAbsoluteCeiling{}, while every world uses the same global vocabulary of 20 directed relations.
Dimensionality and coordinate meaning are therefore fixed across worlds, enabling matched comparison of the two representations across decoders.

\begin{figure*}[t]
\centering
\resizebox{0.94\textwidth}{!}{\input{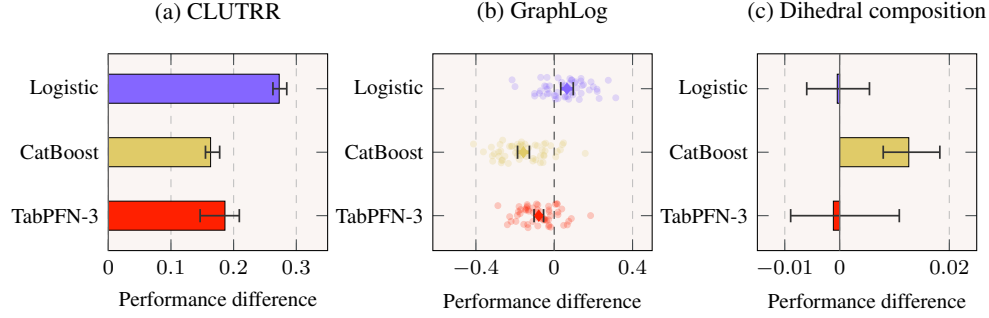}}
\caption{\textbf{Encoder effects across decoders.}
At the largest context budget, values report relation multiset minus
absolute-position performance for (\textit{a}) CLUTRR and (\textit{b})
GraphLog, and coarse minus absolute-position performance for (\textit{c}) the
dihedral control. CLUTRR and the dihedral control use accuracy, while GraphLog
uses macro balanced accuracy. Positive values favour the coarser encoder.
Bars in (\textit{a}) and (\textit{c}) denote means, with capped intervals spanning
paired context seeds and generated populations, respectively.
In (\textit{b}), light points denote paired worlds, diamonds denote means, and
capped intervals denote sensitivity intervals.
GraphLog world-level wins/ties/losses are 41/0/16 for logistic regression,
6/0/51 for \CatBoost{}, and 11/0/46 for \TabPFNthree{}.
Panels use separate horizontal scales.}
\label{fig:composition}
\end{figure*}

For \CatBoost{} and \TabPFNthree{}, absolute positions yield higher balanced accuracy than the relation multiset, while the multiset enters a failure regime in many worlds.
At 40 examples per class, the multiset is at or below each world's chance level in 30 and 12 of the 57 worlds for \CatBoost{} and \TabPFNthree{}, compared with 1 and 4 worlds for absolute positions.
To characterise this regime, a \CatBoost{} tree-depth sweep uses all 57 worlds and holds all other settings fixed across depths 1--7.
Evaluation balanced accuracy changes from \GraphLogDepthOneShared{}/\GraphLogDepthOneAbsolute{} at depth one to \GraphLogDepthSevenShared{}/\GraphLogDepthSevenAbsolute{} at depth seven, widening the gap from \GraphLogDepthOneDelta{} to \GraphLogDepthSevenDelta{}.
Over the same sweep, context fit rises from \GraphLogDepthOneContextShared{}/\GraphLogDepthOneContextAbsolute{} to \GraphLogDepthSevenContextShared{}/\GraphLogDepthSevenContextAbsolute{}.
The complete curve in \cref{app:composition} shows progressively lower multiset evaluation performance as tree depth increases, while absolute-position performance remains comparatively stable.
Together, these results show that the GraphLog encoder ordering changes across decoders and that the \CatBoost{} gap is depth-sensitive, driven mainly by multiset degradation.

The dihedral control limits how broadly the CLUTRR ordering can be interpreted.
Its coarse encoder removes sequence order while also merging distinct group elements.
At chance level \DEightChance{}, coarse/absolute accuracies are \DEightLogShared{}/\DEightLogAbsolute{}, \DEightCatShared{}/\DEightCatAbsolute{}, and \DEightTabShared{}/\DEightTabAbsolute{}.
When coarsening also removes task-relevant identity, both encoders therefore remain at or near chance apart from the small \CatBoost{} contrast.

\subsection{Feature Contributions Are Asymmetric and Context-Dependent}
\label{sec:feature-contributions}

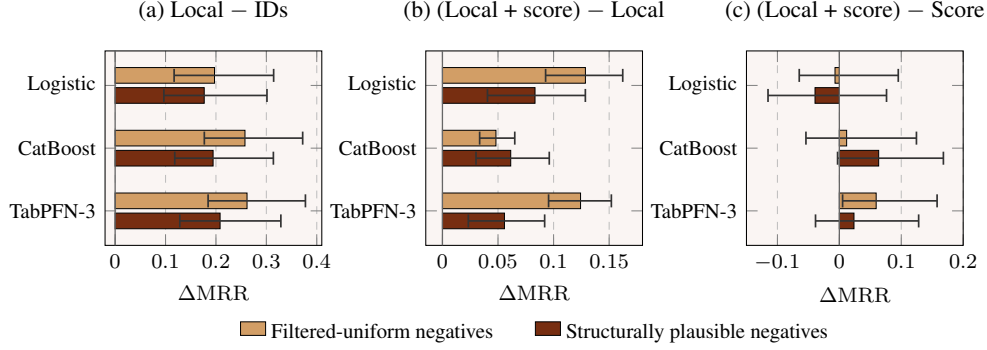
\begin{figure*}[t]
\centering
\resizebox{0.94\textwidth}{!}{\begingroup
\fontfamily{ptm}\selectfont
\definecolor{plotbackground}{HTML}{FAF5F3}%
\definecolor{uniquegamebar}{HTML}{D4AA97}%
\definecolor{nonuniquegamebar}{HTML}{772E11}%
\definecolor{ssl-color}{HTML}{D4AA97}%
\definecolor{sslrl-color}{HTML}{772E11}%
\definecolor{bar-color}{HTML}{D4AA97}%
\definecolor{delta-color}{HTML}{772E11}%
\definecolor{sim-color}{HTML}{D4AA97}%
\definecolor{pos-delta-color}{HTML}{3E9D3A}%
\definecolor{neg-delta-color}{HTML}{EF4444}%
\definecolor{momline}{HTML}{772E11}%
\definecolor{modelsoupline}{HTML}{D29E66}%
\definecolor{transcendenceline}{HTML}{1F65D6}%
\definecolor{karvonenline}{HTML}{CBD77F}%
\definecolor{mixedgmline}{HTML}{99BF31}%
\definecolor{expertarea}{HTML}{729153}%
\definecolor{randompartitionarea}{HTML}{DCC389}%
\definecolor{legendbackground}{HTML}{FFFBF6}%
\colorlet{decoder0}{transcendenceline}%
\colorlet{decoder1}{momline}%
\colorlet{decoder2}{expertarea}%
\begin{tikzpicture}
\begin{groupplot}[group style={group size=3 by 1,horizontal sep=14mm},
width=4.5cm,height=4.2cm,
grid=both,grid style=dashed,axis background/.style={fill=plotbackground},
    every tick label/.append style={font=\fontsize{8}{8}\selectfont},
    scaled ticks=false,tick label style={/pgf/number format/fixed,/pgf/number format/precision=2},
    xlabel style={font=\small},ylabel style={font=\small},title style={font=\small,yshift=2pt},label style={font=\fontsize{8}{8}\selectfont}]
\nextgroupplot[title={(a) Local $-$ IDs},xmin=-0.02,xmax=0.41,ymin=-.55,ymax=2.55,xtick={0,.1,.2,.3,.4},ytick={2,1,0},yticklabels={Logistic,CatBoost,TabPFN-3},xlabel={$\Delta\mathrm{MRR}$},xmajorgrids=true,ymajorgrids=false]
\addplot[xbar,bar shift=0pt,bar width=6pt,fill=modelsoupline,draw=black,line width=.35pt,mark=none,error bars/.cd,x dir=both,x explicit,error bar style={draw=black!80,line width=.65pt},error mark=-,error mark options={rotate=90,draw=black!80,mark size=2.5pt,line width=.65pt}] coordinates {(0.1971330000,2.1600000000) += (0.1173430000,0.0000000000) -= (0.0801350000,0.0000000000) (0.2575910000,1.1600000000) += (0.1144750000,0.0000000000) -= (0.0806470000,0.0000000000) (0.2616840000,0.1600000000) += (0.1158490000,0.0000000000) -= (0.0772480000,0.0000000000)};
\addplot[xbar,bar shift=0pt,bar width=6pt,fill=momline,draw=black,line width=.35pt,mark=none,error bars/.cd,x dir=both,x explicit,error bar style={draw=black!80,line width=.65pt},error mark=-,error mark options={rotate=90,draw=black!80,mark size=2.5pt,line width=.65pt}] coordinates {(0.1767450000,1.8400000000) += (0.1244880000,0.0000000000) -= (0.0800570000,0.0000000000) (0.1943330000,0.8400000000) += (0.1196110000,0.0000000000) -= (0.0760740000,0.0000000000) (0.2082170000,-0.1600000000) += (0.1205880000,0.0000000000) -= (0.0799200000,0.0000000000)};
\draw[black!60,line width=.5pt] (axis cs:0,-0.55) -- (axis cs:0,2.55);
\nextgroupplot[title={(b) (Local + score) $-$ Local},xmin=-0.015,xmax=0.18,ymin=-.55,ymax=2.55,xtick={0,.05,.1,.15},ytick={2,1,0},yticklabels={Logistic,CatBoost,TabPFN-3},xlabel={$\Delta\mathrm{MRR}$},xmajorgrids=true,ymajorgrids=false]
\addplot[xbar,bar shift=0pt,bar width=6pt,fill=modelsoupline,draw=black,line width=.35pt,mark=none,error bars/.cd,x dir=both,x explicit,error bar style={draw=black!80,line width=.65pt},error mark=-,error mark options={rotate=90,draw=black!80,mark size=2.5pt,line width=.65pt}] coordinates {(0.1286340000,2.1600000000) += (0.0335890000,0.0000000000) -= (0.0358080000,0.0000000000) (0.0481400000,1.1600000000) += (0.0169350000,0.0000000000) -= (0.0146600000,0.0000000000) (0.1242670000,0.1600000000) += (0.0277210000,0.0000000000) -= (0.0287640000,0.0000000000)};
\addplot[xbar,bar shift=0pt,bar width=6pt,fill=momline,draw=black,line width=.35pt,mark=none,error bars/.cd,x dir=both,x explicit,error bar style={draw=black!80,line width=.65pt},error mark=-,error mark options={rotate=90,draw=black!80,mark size=2.5pt,line width=.65pt}] coordinates {(0.0832990000,1.8400000000) += (0.0452940000,0.0000000000) -= (0.0427830000,0.0000000000) (0.0614420000,0.8400000000) += (0.0346820000,0.0000000000) -= (0.0312390000,0.0000000000) (0.0558020000,-0.1600000000) += (0.0361650000,0.0000000000) -= (0.0325100000,0.0000000000)};
\draw[black!60,line width=.5pt] (axis cs:0,-0.55) -- (axis cs:0,2.55);
\nextgroupplot[title={(c) (Local + score) $-$ Score},xmin=-0.15,xmax=0.2,ymin=-.55,ymax=2.55,xtick={-.1,0,.1,.2},ytick={2,1,0},yticklabels={Logistic,CatBoost,TabPFN-3},xlabel={$\Delta\mathrm{MRR}$},xmajorgrids=true,ymajorgrids=false]
\addplot[xbar,bar shift=0pt,bar width=6pt,fill=modelsoupline,draw=black,line width=.35pt,mark=none,error bars/.cd,x dir=both,x explicit,error bar style={draw=black!80,line width=.65pt},error mark=-,error mark options={rotate=90,draw=black!80,mark size=2.5pt,line width=.65pt}] coordinates {(-0.0070080000,2.1600000000) += (0.1021770000,0.0000000000) -= (0.0577410000,0.0000000000) (0.0117860000,1.1600000000) += (0.1127180000,0.0000000000) -= (0.0657270000,0.0000000000) (0.0594680000,0.1600000000) += (0.0983980000,0.0000000000) -= (0.0542110000,0.0000000000)};
\addplot[xbar,bar shift=0pt,bar width=6pt,fill=momline,draw=black,line width=.35pt,mark=none,error bars/.cd,x dir=both,x explicit,error bar style={draw=black!80,line width=.65pt},error mark=-,error mark options={rotate=90,draw=black!80,mark size=2.5pt,line width=.65pt}] coordinates {(-0.0393010000,1.8400000000) += (0.1154990000,0.0000000000) -= (0.0758810000,0.0000000000) (0.0635130000,0.8400000000) += (0.1045240000,0.0000000000) -= (0.0663720000,0.0000000000) (0.0236320000,-0.1600000000) += (0.1044390000,0.0000000000) -= (0.0621010000,0.0000000000)};
\draw[black!60,line width=.5pt] (axis cs:0,-0.55) -- (axis cs:0,2.55);
\end{groupplot}\node[anchor=north,font=\fontsize{8}{9}\selectfont,align=center] at ([yshift=-9mm]group c2r1.south) {\tikz[baseline=-.3ex]{\path[fill=modelsoupline,draw=black,line width=.3pt] (0,0) rectangle (1.2em,.65em);} Filtered-uniform negatives\qquad\tikz[baseline=-.3ex]{\path[fill=momline,draw=black,line width=.3pt] (0,0) rectangle (1.2em,.65em);} Structurally plausible negatives};\end{tikzpicture}%
\endgroup}
\caption{\textbf{Paired encoder effects across 57 knowledge graphs.}
Equal-family $\Delta\mathrm{MRR}$ at 200 examples per class for
(\textit{a}) local features minus dataset-scoped entity identifiers,
(\textit{b}) local plus score minus local, and
(\textit{c}) local plus score minus score. Light-brown bars denote filtered-uniform
negatives and dark-brown bars denote structurally plausible negatives. Capped error bars
show 95\% provenance-family bootstrap sensitivity intervals; positive values favour the
first encoder in each contrast. Positive context examples and
evaluation queries are fixed across the two negative-context constructions.}
\label{fig:context_robustness}
\end{figure*}

Across the 57-KG panel, adding the frozen \Ultra{} score to the \emph{Local} encoder improves equal-family mean reciprocal rank (MRR) by \ContextScoreLocalRange{} across all decoder--context combinations.
Five of the six contrasts remain supported after joint correction. The structurally plausible \TabPFNthree{} contrast remains positive but falls just outside the correction threshold.
The reverse addition, adding \emph{Local} features to the score, ranges from \(-0.039\) to \(+0.064\), and none of the six contrasts remains supported after correction.
As a reference comparison, \emph{Local} also outperforms dataset-scoped entity identifiers by \ContextLocalIdsRange{} across all six combinations.
Removing relation identity from \emph{Local} changes equal-family MRR by at most \(0.0031\) in magnitude, and none of the tested contrasts remains supported after joint correction (\cref{app:relation-id}).
\cref{fig:context_robustness} summarises these three matched comparisons.

The native ranking provides a rank-preservation sanity check.
Score-only logistic regression reproduces the native \Ultra{} ranks exactly across all \UltraScoreRankAuditCells{} combinations of dataset, context budget, and seed, as expected when the one-dimensional frozen score is read out with positive orientation.
For \TabPFNthree{}, adding global PageRank, component, heat, and diffusion features after the score changes MRR by only \UltraGlobalGivenScore{} (\UltraGlobalGivenScoreCI{}), with the contrast falling below the joint-correction threshold.
\TabPFNthree{} with the \emph{Full} encoder underperforms native \Ultra{} on 48 of 57 datasets and 14 of 16 provenance families.
Its full-minus-native contrast is \UltraFullNative{} under equal-family weighting but \UltraFullNativeDatasetMean{} under equal-dataset weighting. The corresponding medians are \UltraFullNativeFamilyMedian{} and \UltraFullNativeDatasetMedian{}.
The leave-one-family-out range is \UltraFullNativeLOFO{}, with the minimum obtained by removing MetaFam.
The change between equal-family and equal-dataset aggregation shows that this panel-level comparison is sensitive to weighting.

Because the tabular decoders are pointwise while evaluation ranks candidates within each query, we test their sensitivity to label-free within-query z-score and percentile transformations.
These improve local-only logistic regression by \QueryRelLogLocalZDelta{}/\QueryRelLogLocalPctDelta{} MRR, but yield negative point estimates for every tested \CatBoost{} and \TabPFNthree{} transformation.
All six \CatBoost{} reductions and five of the six \TabPFNthree{} reductions remain supported after joint correction.
Score-only logistic regression preserves all \QueryRelRanks{} native query ranks exactly under both maps.
Query-relative scaling therefore affects the tested pointwise readouts in a decoder-specific way. The analysis does not alter the underlying ranking objective (\cref{app:query-relative}).
A 512-query sensitivity preserves the sign of all five panel-level encoder contrasts, with mean changes no larger than 0.0064 MRR (\cref{app:ultra57}).

Changing only the negative context also affects performance.
Structurally plausible negatives reduce MRR by \ContextInformativeDropRange{} across all nine informative encoder--decoder combinations, whereas identifier baselines change by less than 0.001.
The selector uses only graph-based criteria, yet its selected negatives have a mean frozen score of \(-1.238\), compared with \(-6.929\) under filtered-uniform sampling, consistent with a harder negative distribution under frozen \Ultra{}.
Decoder ordering also varies with the encoder: logistic regression is strongest for the score-only row, whereas \TabPFNthree{} is strongest for the local and local-plus-score rows (\cref{app:ultra57}).

A separate matched control isolates all-known-positive rejection from the other differences between the two context constructions.
Filtering changes equal-family MRR by \FilterLocalFamilyDelta{} for \emph{Local}, \FilterLocalScoreFamilyDelta{} for \emph{Local + score}, and \FilterScoreFamilyDelta{} for score-only \TabPFNthree{}, with all three contrasts below the joint-correction threshold.
The matched control therefore indicates that filtering accounts for only a small part of the differences between the two context constructions (\cref{app:negative-context}).
We also vary positive-context selection.
Matching positives to the evaluation relation distribution increases relation-type coverage from \(0.667\) to \(0.871\) and relation mass from \(0.693\) to \(0.891\), while reducing distributional total variation from \(0.447\) to \(0.190\).
The corresponding MRR effects are generally small. Only \CatBoost{} with the \emph{Local} encoder without relation identity remains supported after joint correction (\(+0.0139\); \cref{app:relation-context}).

\subsection{Local Feature Utility Varies with Endpoint Connectivity}
\label{sec:endpoint-connectivity}

Aggregate MRR conceals substantial differences among query regimes.
We stratify the 7,296 directional queries by observed endpoint connectivity in the support graph, excluding the held-out target edge. 1,622 have a direct or reverse endpoint-pair edge.

Appendix \cref{fig:query_evidence} shows that dyadic features achieve higher aggregate reciprocal rank than bounded-path features in each endpoint-connected stratum.
The ordering reverses for the \QueryNeither{} queries without a pair edge.
Dyadic reciprocal rank falls to \QueryDyadicNeitherRange{}, whereas path features reach \QueryPathsNeitherRange{} and the \emph{Local} encoder reaches \QueryLocalNeitherRange{}.
The broader \emph{Local} encoder therefore retains higher ranking performance for the majority of queries where direct pair evidence is absent.

The both-direction and direct-only groups contain \QueryBoth{} and \QueryDirect{} queries, so we treat their absolute levels as fixed-panel descriptions.
Complete counts are reported in \cref{app:query-evidence}, and the descriptive, order-dependent variance decomposition in \cref{app:screening}.
The stratification therefore identifies query regimes in which different graph-derived feature groups are associated with higher ranking performance.
\section{Conclusion}

We introduced a paired diagnostic framework for the graph-to-table encoder \(\gttphi\), context constructor \(\gttS\), and decoder \(\gttD\).
On CLUTRR, preserving query distinctions is insufficient for length transfer: unsupported absolute coordinates have negligible influence, while the aligned representation achieves near-perfect accuracy on the labelled context rows yet transfers worse to longer paths.
On GraphLog, encoder preference changes across decoders, and a \CatBoost{} tree-depth sweep shows that the gap between absolute positions and relation counts widens primarily through degradation of the latter.
Across 57 knowledge graphs, adding a frozen \Ultra{} score to local features provides consistent gains, whereas adding local features to the score has a smaller and less stable effect.
A pointwise \TabPFNthree{} readout with the \emph{Full} encoder underperforms native \Ultra{} on 48 of 57 datasets, while its aggregate comparison changes from positive under equal-family weighting to negative under equal-dataset weighting.
Changing the labelled context also changes the measured contribution of the same encoder.
Together, these results show that graph-to-table representations cannot be assessed through the encoder alone.
Their behaviour depends jointly on the labelled context, decoder, evaluation protocol, and panel aggregation.

\clearpage
\bibliographystyle{plainnat}
\bibliography{references}

\clearpage
\appendix
\crefalias{section}{appendix}
\crefalias{subsection}{subappendix}
\crefalias{subsubsection}{subsubappendix}
\section{Limitations}

Our experiments audit the graph-to-table interface alone. CLUTRR and GraphLog use supplied symbolic paths, so they evaluate encoding and decoding after relational evidence has been extracted. The observed encoder, context, and decoder effects are limited to the tested configurations, while the 57-KG study uses constructed feature groups and one frozen upstream checkpoint whose pretraining lineage overlaps with part of the panel. Query-level groups and the sequential variance decomposition are descriptive and do not establish causal feature importance or a deployable routing rule. Finally, the family-weighted KG panel uses 64 sampled test triples per dataset. All-entity filtered ranking and the 512-query sensitivity support the reported contrast directions, but the reported sensitivities are conditional on this fixed panel.

\section{Controlled Calibration of the Audit Diagnostics}
\label{app:controlled}

This appendix provides the complete protocols, robustness analyses, and reproducibility details supporting \cref{sec:results}. The controlled calibration first tests whether the diagnostics in \cref{sec:audit} recover known properties under graph-disjoint evaluation, before applying them to the benchmark-specific analyses in \cref{sec:results}. We generate four binary tasks governed by reciprocity, typed relation composition, local structural roles, and shared graph-community membership. For each task, ten independently generated graphs define disjoint context and evaluation populations. We compute the latent-label ceiling before applying a deterministic 10\% symmetric label flip, and define mixed-label collision mass as the fraction of examples assigned to a row-equivalence class containing both latent labels.

\subsection{Finite-Population Diagnostics}

\cref{fig:controlled_ladder} reports fitted ROC-AUC (area under the receiver operating characteristic curve) after the controlled label perturbation, whereas the finite-population diagnostics below are decoder-independent properties of the encoder.

We first verify that the ceiling, singleton-mass, and collision diagnostics recover the distinctions imposed by each generated property. For row-equivalence class \(c\), let \(n_c^+\) and \(n_c^-\) denote its latent positive and negative counts. We define mixed-label collision mass as
\begin{equation}
\kappa_{\gttphi}=\frac{1}{N}\sum_c(n_c^++n_c^-)\mathbf{1}[n_c^+>0]\mathbf{1}[n_c^->0],
\label{eq:collision-mass}
\end{equation}
where \(N=\sum_c(n_c^++n_c^-)\). \cref{eq:collision-mass} is the fraction of examples assigned to a row-equivalence class containing both latent labels.

\begin{table}[!ht]
\centering
\scriptsize
\caption{\textbf{Finite-population row-equivalence diagnostics.}
Results are averaged over ten independently generated evaluation graphs. The
ceiling and collision mass use latent labels before the deterministic 10\%
symmetric label flip. A ceiling and singleton mass of one indicate that every
evaluated query has a unique row, characterising distinguishability on the evaluated population.}
\label{tab:controlled}
\begin{adjustbox}{max width=\linewidth}
\begin{tabular}{lrrr}
\toprule
Encoder / property & Latent \(\Astar\) & Singleton mass & Collision mass \\
\midrule
Relation only / reciprocity & \ControlRelationCeiling{} & 0.000 & \ControlRelationCollision{} \\
Reverse edge / reciprocity & \ControlReciprocityCeiling{} & \ControlReciprocitySingleton{} & 0.000 \\
Typed paths / composition & \ControlCompositionCeiling{} & \ControlCompositionSingleton{} & 0.000 \\
Local roles / structural role & \ControlRolesCeiling{} & \ControlRolesSingleton{} & 0.000 \\
Community affinity / community & \ControlBlockCeiling{} & 0.655 & \ControlBlockCollision{} \\
\bottomrule
\end{tabular}
\end{adjustbox}
\end{table}

For binary ranking, the finite-population AUC ceiling follows the optimal ordering of row-equivalence classes by posterior positive probability~\citep{menon2016bipartite}. For class \(c\), let \(n_c^+\), \(n_c^-\), and \(\eta_c=n_c^+/(n_c^++n_c^-)\) denote its positive count, negative count, and positive rate. With \(N^+=\sum_c n_c^+\) and \(N^-=\sum_c n_c^-\), the ceiling is
\begin{equation}
\mathrm{AUC}^{\star}(\gttphi)=\frac{\sum_{c,d}n_c^+n_d^-\left[\mathbf 1(\eta_c>\eta_d)+\tfrac12\mathbf 1(\eta_c=\eta_d)\right]}{N^+N^-}.
\label{eq:auc-ceiling}
\end{equation}
Ordering classes by \(\eta_c\) is optimal because swapping an inverted pair increases the number of correctly ordered positive--negative pairs. \cref{eq:auc-ceiling}, like the accuracy ceiling in \cref{sec:audit}, is an oracle diagnostic on a fixed labelled population.

Each encoder begins with a four-way query-relation one-hot vector. The reciprocity encoder adds direct- and reverse-edge indicators and endpoint in- and out-degree. The typed-composition encoder adds path counts and existence indicators for fixed ordered two-relation templates. The structural-role encoder adds relation-specific endpoint degrees. The community-affinity encoder adds endpoint-degree contrasts, common-neighbour count, Jaccard similarity, Adamic--Adar, preferential attachment, and component status. Feature construction is independent of the perturbed target label.

The calibration flips every tenth positive and negative candidate after an independent within-class permutation determined by the split seed. Query index, candidate order, and flip status are not encoded. Graph generation, relabelling, splitting, feature construction, noise, and decoder seeds are distinct. Because the perturbation mechanically caps sufficient repeated-row encoders near 0.9, \cref{tab:controlled} reports latent pre-flip ceilings and collisions for interpreting encoder distinguishability.

\subsection{Encoder Ladder and Alignment Control}

\begin{figure}[!ht]
\centering
\resizebox{\linewidth}{!}{\input{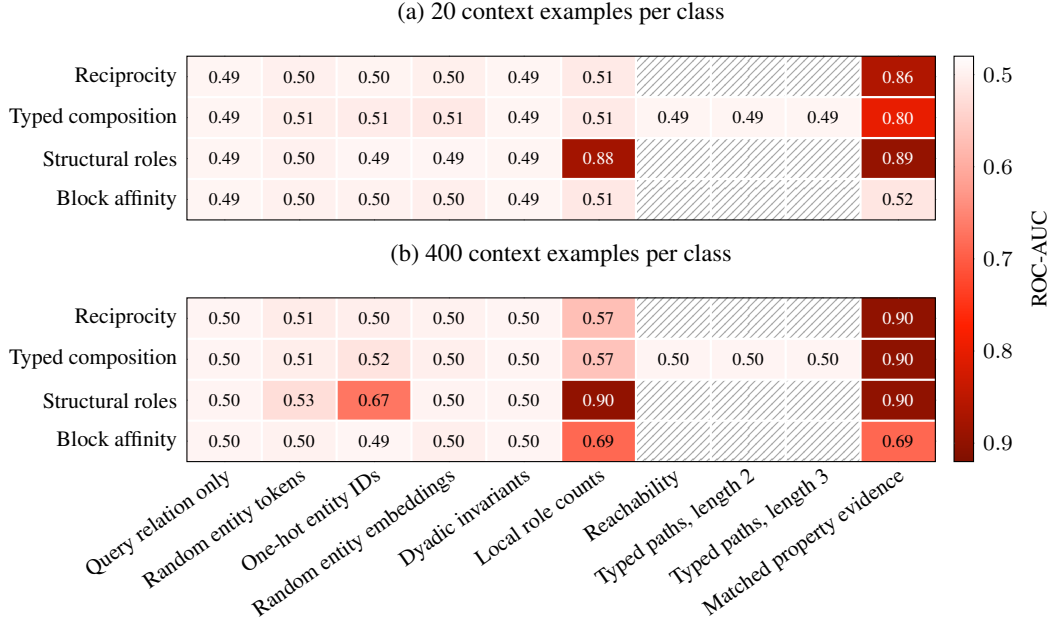}}
\caption{\textbf{Graph-disjoint encoder ladder.}
Frozen \textsc{TabPFN-3} ROC-AUC for the four generated relational properties
(rows) and candidate encoders (columns). Panel (\textit{a}) uses 20
labelled context examples per class and panel (\textit{b}) uses 400. Each cell
reports the mean over ten graph seeds, and both panels share the same colour scale.
Matched property evidence is the aligned evidence bank constructed for each property.
Hatched cells denote encoders not instantiated for the corresponding property.}
\label{fig:controlled_ladder}
\end{figure}

We next test whether the encoder ladder and row-permutation control distinguish task-specific evidence from dimension-matched but misaligned features. \cref{fig:controlled_ladder} reports that the constructed evidence is property-specific. Local-role counts solve the structural-role task but remain near chance on reciprocity and typed composition, which require reverse-edge and ordered-path evidence. The off-diagonal cells therefore test whether a feature family generalises beyond its construction.

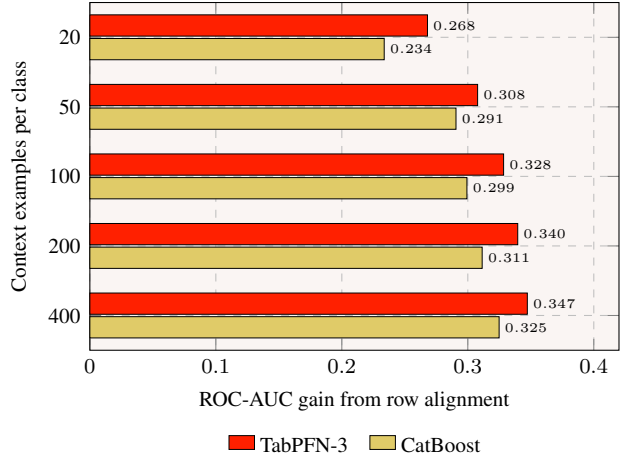
\begin{figure}[!ht]
\centering
\begingroup
\fontfamily{ptm}\selectfont
\definecolor{plotbackground}{HTML}{FAF5F3}%
\colorlet{decoder0}{blue!80!magenta!60!white}%
\definecolor{decoder1}{HTML}{E0CB68}%
\colorlet{decoder2}{red!75!orange}%
\begin{tikzpicture}
\begin{axis}[scale only axis,width=7cm,height=4.6cm,clip=false,y dir=reverse,xmin=0,xmax=.42,xtick={0,.1,.2,.3,.4},
ymin=-.5,ymax=4.5,ytick={0,...,4},yticklabels={20,50,100,200,400},xmajorgrids=true,ymajorgrids=false,grid=both,
xlabel={ROC-AUC gain from row alignment},ylabel={Context examples per class},
grid style=dashed,axis background/.style={fill=plotbackground},
    every tick label/.append style={font=\fontsize{8}{8}\selectfont},
    scaled ticks=false,tick label style={/pgf/number format/fixed,/pgf/number format/precision=2},
    xlabel style={font=\small},ylabel style={font=\small},title style={font=\small,yshift=2pt},label style={font=\fontsize{8}{8}\selectfont}]
\addplot[xbar,bar shift=0pt,bar width=8pt,fill=decoder2,draw=black,line width=.35pt] coordinates {(0.2679864844,-0.17) (0.3077773437,0.83) (0.3284054687,1.83) (0.3396260156,2.83) (0.3472436719,3.83)};
\node[anchor=west,inner sep=0pt,font=\fontsize{5}{5}\selectfont,text=black] at ([xshift=2pt]axis cs:0.26798648437500006,-0.17) {\pgfmathprintnumber[fixed,fixed zerofill,precision=3]{0.2679864844}};
\node[anchor=west,inner sep=0pt,font=\fontsize{5}{5}\selectfont,text=black] at ([xshift=2pt]axis cs:0.30777734374999993,0.83) {\pgfmathprintnumber[fixed,fixed zerofill,precision=3]{0.3077773437}};
\node[anchor=west,inner sep=0pt,font=\fontsize{5}{5}\selectfont,text=black] at ([xshift=2pt]axis cs:0.32840546875,1.83) {\pgfmathprintnumber[fixed,fixed zerofill,precision=3]{0.3284054687}};
\node[anchor=west,inner sep=0pt,font=\fontsize{5}{5}\selectfont,text=black] at ([xshift=2pt]axis cs:0.33962601562500006,2.83) {\pgfmathprintnumber[fixed,fixed zerofill,precision=3]{0.3396260156}};
\node[anchor=west,inner sep=0pt,font=\fontsize{5}{5}\selectfont,text=black] at ([xshift=2pt]axis cs:0.34724367187500005,3.83) {\pgfmathprintnumber[fixed,fixed zerofill,precision=3]{0.3472436719}};
\addplot[xbar,bar shift=0pt,bar width=8pt,fill=decoder1,draw=black,line width=.35pt] coordinates {(0.2336402344,0.17) (0.2906195312,1.17) (0.2992991406,2.17) (0.3113983594,3.17) (0.3248542187,4.17)};
\node[anchor=west,inner sep=0pt,font=\fontsize{5}{5}\selectfont,text=black] at ([xshift=2pt]axis cs:0.23364023437500003,0.17) {\pgfmathprintnumber[fixed,fixed zerofill,precision=3]{0.2336402344}};
\node[anchor=west,inner sep=0pt,font=\fontsize{5}{5}\selectfont,text=black] at ([xshift=2pt]axis cs:0.29061953124999995,1.17) {\pgfmathprintnumber[fixed,fixed zerofill,precision=3]{0.2906195312}};
\node[anchor=west,inner sep=0pt,font=\fontsize{5}{5}\selectfont,text=black] at ([xshift=2pt]axis cs:0.29929914062499996,2.17) {\pgfmathprintnumber[fixed,fixed zerofill,precision=3]{0.2992991406}};
\node[anchor=west,inner sep=0pt,font=\fontsize{5}{5}\selectfont,text=black] at ([xshift=2pt]axis cs:0.31139835937499993,3.17) {\pgfmathprintnumber[fixed,fixed zerofill,precision=3]{0.3113983594}};
\node[anchor=west,inner sep=0pt,font=\fontsize{5}{5}\selectfont,text=black] at ([xshift=2pt]axis cs:0.32485421875,4.17) {\pgfmathprintnumber[fixed,fixed zerofill,precision=3]{0.3248542187}};
\end{axis}\node[anchor=north,font=\fontsize{8}{9}\selectfont,align=center] at ([yshift=-10mm]current axis.south) {\tikz[baseline=-.3ex]{\path[fill=decoder2,draw=black,line width=.3pt] (0,0) rectangle (1.2em,.65em);} TabPFN-3\quad\tikz[baseline=-.3ex]{\path[fill=decoder1,draw=black,line width=.3pt] (0,0) rectangle (1.2em,.65em);} CatBoost};
\end{tikzpicture}%
\endgroup
\caption{\textbf{Effect of row alignment across context budgets.}
Bars report the equal-property macro ROC-AUC gain of correctly aligned rows over a joint
row permutation of the complete feature block, for \textsc{TabPFN-3} and \textsc{CatBoost}
at each number of labelled context examples per class, averaged over ten regenerated graphs
per property. Under the permutation, ROC-AUC stays between 0.497 and 0.508, close to
chance (0.5). Bar labels are rounded to three decimals.}
\label{fig:controlled_context}
\end{figure}

\cref{fig:controlled_context} compares aligned rows with a joint row-permutation null that preserves dimensionality and the covariance structure of complete feature blocks while breaking query-to-row correspondence. Aligned performance increases with the context budget, whereas both permuted conditions remain near chance. At 20 examples per class, the pre-specified \TabPFNthree{}--\CatBoost{} difference-in-differences is \(+0.0343\), with fixed-panel sensitivity \([-0.0128,+0.1106]\). The calibration therefore validates the alignment control.

A separately audited independent-column permutation preserves the marginal distribution of each aligned column while destroying cross-column covariance and yields the same near-chance conclusion. Because either permutation can also alter dependence between the permuted block and unpermuted columns, we interpret these variants as broad alignment nulls.

\section{Composition Benchmarks and Complete Results}
\label{app:composition}

This section provides the complete encoder schemas, absolute performance levels, and robustness checks for the compositional results in \cref{sec:clutrr} and \cref{sec:decoder-dependence}. The ordering follows the main analysis. We first document the CLUTRR alignment and length-exposure controls, then report the GraphLog decoder sensitivities and the dihedral boundary control.

\subsection{CLUTRR Protocol and Encoder Schemas}
\label{app:clutrr}

The official \textit{gen-train23-test2to10} split contains 9,074 training, 2,020 validation, and 1,146 test stories. Its 18 target classes, in configured index order, are \{\texttt{aunt}, \texttt{son-in-law}, \texttt{grandfather}, \texttt{brother}, \texttt{sister}, \texttt{father}, \texttt{mother}, \texttt{grandmother}, \texttt{uncle}, \texttt{daughter-in-law}, \texttt{grandson}, \texttt{granddaughter}, \texttt{father-in-law}, \texttt{mother-in-law}, \texttt{nephew}, \texttt{son}, \texttt{daughter}, \texttt{niece}\}. The path-edge vocabulary is the union over the fixed train, validation, and test files and contains \{\texttt{aunt}, \texttt{brother}, \texttt{daughter}, \texttt{father}, \texttt{granddaughter}, \texttt{grandfather}, \texttt{grandmother}, \texttt{grandson}, \texttt{husband}, \texttt{mother}, \texttt{sister}, \texttt{son}, \texttt{uncle}, \texttt{wife}\}.

\begin{table}[!ht]
\centering
\scriptsize
\caption{\textbf{CLUTRR encoder schemas.}
All encoders use the same 14-symbol path-edge vocabulary and exclude
entity identifiers. Combined variants retain the 15-column relation-multiset
block and differ only in the treatment of the 140-column absolute-position
block.}
\label{tab:clutrr_schemas}
\begin{adjustbox}{max width=\linewidth}
\begin{tabular}{lrl}
\toprule
Encoder & Dim. & Contents \\
\midrule
Relation multiset & 15 & Path length and one count per relation symbol \\
Absolute position & 141 & Path length and ten 14-way positional one-hot blocks \\
Directed transitions & 197 & Path length and counts of ordered adjacent relation pairs \\
Aligned absolute + multiset & 155 & Relation multiset and the aligned absolute-position block \\
Suffix-zeroed absolute + multiset & 155 & Aligned positions 1--3, with positions 4--10 set to zero \\
Permuted absolute + multiset & 155 & Relation multiset and a complete absolute block reassigned within length \\
\bottomrule
\end{tabular}
\end{adjustbox}
\end{table}

The permutation acts on complete absolute-position rows separately within each path length. For each paired context seed \(s\), deterministic within-length permutations
are generated separately for context and evaluation using seeds \(s\) and
\(s+1\), respectively. This preserves dimensionality and the empirical joint distribution of complete absolute-position blocks within each length while breaking their assignment to the original paths. All decoders receive identical stories, labels, context indices, and evaluation queries.

\subsection{CLUTRR Complete Results and Robustness}

\cref{tab:clutrr_levels} reports the absolute levels underlying \cref{fig:clutrr}. At 80 examples per target relation, the relation-multiset encoder outperforms the absolute-position encoder for all three decoders. The directed-transition encoder retains local order without absolute positions, but its relative performance changes across decoders.

\begin{table}[!ht]
\centering
\scriptsize
\caption{\textbf{CLUTRR OOD accuracy.}
Results use 80 context examples per target relation and are averaged equally
across evaluation lengths 4--10 and then across the three paired context seeds.
Values in parentheses denote encoder dimensionality.}
\label{tab:clutrr_levels}
\setlength{\tabcolsep}{5pt}
\begin{adjustbox}{max width=\linewidth}
\begin{tabular}{lrrr}
\toprule
Decoder & Multiset (15) & Transitions (197) & Absolute (141) \\
\midrule
Logistic & \ClutrrLogMultiset{} & \ClutrrLogTransition{} & \ClutrrLogAbsolute{} \\
\CatBoost{} & \ClutrrCatMultiset{} & \ClutrrCatTransition{} & \ClutrrCatAbsolute{} \\
\TabPFNthree{} & \ClutrrTabMultiset{} & \ClutrrTabTransition{} & \ClutrrTabAbsolute{} \\
\bottomrule
\end{tabular}
\end{adjustbox}

\medskip
\begin{adjustbox}{max width=\linewidth}
\begin{tabular}{lrrr}
\toprule
Decoder & Aligned + multiset & Suffix-zeroed & Permuted + multiset \\
& (155) & (155) & (155) \\
\midrule
Logistic & \ClutrrLogCombined{} & \ClutrrLogCombined{} & \ClutrrLogPermuted{} \\
\CatBoost{} & \ClutrrCatCombined{} & \ClutrrCatCombined{} & \ClutrrCatPermuted{} \\
\TabPFNthree{} & \ClutrrTabCombined{} & \ClutrrTabCombined{} & \ClutrrTabPermuted{} \\
\bottomrule
\end{tabular}
\end{adjustbox}
\end{table}

The suffix-zeroed and fully aligned encoders yield indistinguishable OOD accuracy, with a maximum paired accuracy change of \(2.8\times10^{-17}\). The suffix-zeroed ceiling is \ClutrrPrefixCeiling{} with singleton mass \ClutrrPrefixSingleton{}, so only \ClutrrPrefixNonSingletonMass{} of the equal-length evaluation mass contributes repeated rows. Its conditional repeated-row ceiling is \ClutrrPrefixNonSingletonCeiling{}. The absolute-position and combined encoders fit the length-2--3 context at \ClutrrContextFitRange{} accuracy, as reported in \cref{tab:clutrr_context_fit}.

\begin{table}[!ht]
\centering
\scriptsize
\caption{\textbf{CLUTRR context fit and finite-population diagnostics.}
Results use 80 context examples per target relation. OOD diagnostics average
lengths 4--10 equally. Conditional ceilings use only repeated-row mass and are
undefined when every row is a singleton. Context accuracy is averaged across
the same three paired context seeds used for OOD evaluation.}
\label{tab:clutrr_context_fit}
\begin{adjustbox}{max width=\linewidth}
\begin{tabular}{lrrrrrrr}
\toprule
Encoder & OOD ceiling & Singleton & Non-singleton & Conditional & Logistic & \CatBoost{} & \TabPFNthree{} \\
\midrule
Relation multiset & \ClutrrMultisetCeiling{} & \ClutrrMultisetSingleton{} & \ClutrrMultisetNonSingletonMass{} & \ClutrrMultisetNonSingletonCeiling{} & 0.764 & 0.774 & 0.791 \\
Absolute position & \ClutrrAbsoluteCeiling{} & \ClutrrAbsoluteSingleton{} & 0.000 & -- & 0.990 & 0.991 & 0.991 \\
Supported prefix + multiset & \ClutrrPrefixCeiling{} & \ClutrrPrefixSingleton{} & \ClutrrPrefixNonSingletonMass{} & \ClutrrPrefixNonSingletonCeiling{} & 0.994 & 0.995 & 0.995 \\
\bottomrule
\end{tabular}
\end{adjustbox}
\end{table}

The collision structure also changes with path length. The relation-multiset singleton mass increases from \(0.453\) at length~4 to \(0.933\) at length~10, whereas the absolute-position encoder has singleton mass \(1.000\) at every evaluated length. High finite-population ceilings therefore increasingly reflect query-level lookup as paths become longer.

\cref{tab:clutrr_interventions} separates block addition, block alignment, and length exposure. The aligned and permuted conditions differ only in path-to-block assignment. The exposure intervention retains the same nested length-2--3 base in both arms and adds 20 examples per target relation from either lengths 2--3 or length 4.

\begin{table}[!ht]
\centering
\scriptsize
\caption{\textbf{Matched CLUTRR interventions.}
Block addition reports permuted absolute plus multiset minus multiset only,
alignment reports permuted absolute plus multiset minus aligned absolute plus
multiset, and length exposure replaces 20 additional length-2--3 examples with
20 length-4 examples. Positive values denote higher OOD accuracy for block
addition and length exposure. For alignment, they denote the accuracy lost
after restoring path-level alignment.}
\label{tab:clutrr_interventions}
\begin{adjustbox}{max width=\linewidth}
\begin{tabular}{lrrrr}
\toprule
Decoder & Block addition & Alignment & Exposure at L4 & Exposure at L5--10 \\
\midrule
Logistic & -0.030 & \ClutrrLogAlignCost{} & \ClutrrAddLogFour{} & \ClutrrAddLogLong{} \\
\CatBoost{} & -0.017 & \ClutrrCatAlignCost{} & \ClutrrAddCatFour{} & \ClutrrAddCatLong{} \\
\TabPFNthree{} & +0.002 & \ClutrrTabAlignCost{} & \ClutrrAddTabFour{} & \ClutrrAddTabLong{} \\
\bottomrule
\end{tabular}
\end{adjustbox}
\end{table}

\cref{tab:clutrr_metric_robustness} shows that the multiset-minus-absolute ordering holds across every declared classification endpoint. For log loss and Brier score, positive values denote the reduction obtained by the relation multiset.

\begin{table}[!ht]
\centering
\scriptsize
\caption{\textbf{CLUTRR metric robustness.}
Results use 80 context examples per target relation and are averaged equally
across OOD lengths 4--10 and then across the three paired context seeds.
Positive values favour the relation multiset.}
\label{tab:clutrr_metric_robustness}
\setlength{\tabcolsep}{3pt}
\begin{adjustbox}{max width=\linewidth}
\begin{tabular}{lrrrrrr}
\toprule
Decoder & Accuracy & Bal. acc. & Macro-F1 & Top-3 & Log loss & Brier \\
\midrule
Logistic & +0.273 & +0.236 & +0.224 & +0.403 & +4.208 & +0.482 \\
\CatBoost{} & +0.163 & +0.132 & +0.155 & +0.293 & +2.002 & +0.459 \\
\TabPFNthree{} & +0.186 & +0.201 & +0.153 & +0.304 & +2.510 & +0.540 \\
\bottomrule
\end{tabular}
\end{adjustbox}
\end{table}

The relation multiset outperforms absolute positions on 16 of 18 target relations for logistic regression and \CatBoost{}, and on 15 of 18 for \TabPFNthree{}. Restricting evaluation to lengths 4--5, matching the magnitude of the GraphLog length shift, retains effects of \(+0.343/+0.177/+0.338\). These analyses are robustness checks within one official population. The shared-parameter GRU uses a 32-dimensional relation embedding, 64 hidden units, one recurrent layer, and context-only early stopping. Its accuracy increases from 0.346 to 0.705 across the five context budgets and decreases from 0.898 at length 4 to 0.527 at length 10 at the largest budget. These controls support the separation in \cref{sec:clutrr} between row distinguishability, context fit, and short-to-long transfer.

\subsection{GraphLog Complete Results and Decoder Sensitivities}

All 57 GraphLog worlds use the same global vocabulary of 20 directed relation symbols. Encoder dimensionality and coordinate meaning therefore remain fixed across worlds. \cref{tab:composition_levels} provides the absolute balanced-accuracy levels behind \cref{fig:composition}, together with the absolute levels for the dihedral boundary control.

\begin{table}[!ht]
\centering
\scriptsize
\caption{\textbf{GraphLog and dihedral-control results.}
GraphLog reports equal-world macro balanced accuracy at the largest context
budget, with the contrast defined as relation multiset minus absolute position.
The dihedral control reports accuracy over the 16 symmetries of a regular
octagon and has chance accuracy $1/16=0.0625$.}
\label{tab:composition_levels}
\begin{adjustbox}{max width=\linewidth}
\begin{tabular}{lrrrrr}
\toprule
& \multicolumn{3}{c}{GraphLog} & \multicolumn{2}{c}{Dihedral control} \\
\cmidrule(lr){2-4}\cmidrule(l){5-6}
Decoder & Multiset & Absolute & Difference & Coarse & Absolute \\
\midrule
Logistic & \GraphLogLogShared{} & \GraphLogLogAbsolute{} & \GraphLogLogDelta{} & \DEightLogShared{} & \DEightLogAbsolute{} \\
\CatBoost{} & \GraphLogCatShared{} & \GraphLogCatAbsolute{} & \GraphLogCatDelta{} & \DEightCatShared{} & \DEightCatAbsolute{} \\
\TabPFNthree{} & \GraphLogTabShared{} & \GraphLogTabAbsolute{} & \GraphLogTabDelta{} & \DEightTabShared{} & \DEightTabAbsolute{} \\
\bottomrule
\end{tabular}
\end{adjustbox}
\end{table}

The multiset--absolute comparison is complemented by a blockwise alignment control. At 40 examples per class, the block-permuted-minus-aligned contrast is small for logistic regression, whereas both nonlinear decoders prefer the correctly aligned ordered block. Because the permutation removes both supported prefix coordinates and evaluation-only suffix coordinates, this contrast measures broad ordered-block dependence.

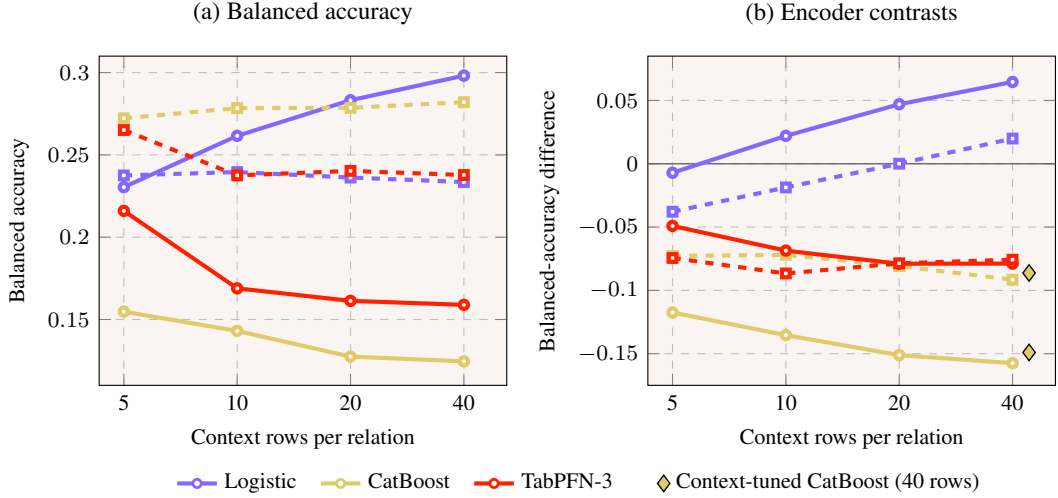
\begin{figure}[!ht]
\centering
\resizebox{\linewidth}{!}{\begingroup
\fontfamily{ptm}\selectfont
\definecolor{plotbackground}{HTML}{FAF5F3}%
\colorlet{decoder0}{blue!80!magenta!60!white}%
\definecolor{decoder1}{HTML}{E0CB68}%
\colorlet{decoder2}{red!75!orange}%
\begin{tikzpicture}
\begin{groupplot}[group style={group size=2 by 1,horizontal sep=18mm},
scale only axis,width=5.2cm,height=4.2cm,xmode=log,log basis x=2,xmin=4.3,xmax=52,
xtick={5,10,20,40},xticklabels={5,10,20,40},xminorticks=false,xmajorgrids=true,ymajorgrids=true,
xlabel={Context rows per relation},
grid style=dashed,axis background/.style={fill=plotbackground},
    every tick label/.append style={font=\fontsize{8}{8}\selectfont},
    scaled ticks=false,tick label style={/pgf/number format/fixed,/pgf/number format/precision=2},
    xlabel style={font=\small},ylabel style={font=\small},title style={font=\small,yshift=2pt},label style={font=\fontsize{8}{8}\selectfont}]
\nextgroupplot[title={(a) Balanced accuracy},ymin=.11,ymax=.31,ytick={.15,.2,.25,.3},ylabel={Balanced accuracy}]
\addplot[color=decoder0,line width=1.5pt,mark=*,mark size=1.5pt,mark options={fill=white,solid}] coordinates {(5,0.2303969193) (10,0.2615740000) (20,0.2832097681) (40,0.2981368725)};
\addplot[color=decoder0,line width=1.5pt,dashed,mark=square*,mark size=1.4pt,mark options={fill=white,solid}] coordinates {(5,0.2375181145) (10,0.2395573063) (20,0.2362242983) (40,0.2335369112)};
\addplot[color=decoder1,line width=1.5pt,mark=*,mark size=1.5pt,mark options={fill=white,solid}] coordinates {(5,0.1547838810) (10,0.1430677512) (20,0.1274636203) (40,0.1245593911)};
\addplot[color=decoder1,line width=1.5pt,dashed,mark=square*,mark size=1.4pt,mark options={fill=white,solid}] coordinates {(5,0.2722249063) (10,0.2784244846) (20,0.2785735656) (40,0.2820264297)};
\addplot[color=decoder2,line width=1.5pt,mark=*,mark size=1.5pt,mark options={fill=white,solid}] coordinates {(5,0.2160153098) (10,0.1689267080) (20,0.1613346094) (40,0.1589084188)};
\addplot[color=decoder2,line width=1.5pt,dashed,mark=square*,mark size=1.4pt,mark options={fill=white,solid}] coordinates {(5,0.2651098597) (10,0.2375353400) (20,0.2402751867) (40,0.2377804368)};
\nextgroupplot[title={(b) Encoder contrasts},ymin=-.175,ymax=.085,ytick={-.15,-.1,-.05,0,.05},ylabel={Balanced-accuracy difference}]
\addplot[black!60,forget plot,no marks,line width=.5pt] coordinates {(4.3,0) (52,0)};
\addplot[color=decoder0,line width=1.5pt,mark=*,mark size=1.5pt,mark options={fill=white,solid}] coordinates {(5,-0.0071211952) (10,0.0220166937) (20,0.0469854698) (40,0.0645999613)};
\addplot[color=decoder0,line width=1.5pt,dashed,mark=square*,mark size=1.4pt,mark options={fill=white,solid}] coordinates {(5,-0.0379085179) (10,-0.0187076044) (20,0.0000504653) (40,0.0198951794)};
\addplot[color=decoder1,line width=1.5pt,mark=*,mark size=1.5pt,mark options={fill=white,solid}] coordinates {(5,-0.1174410252) (10,-0.1353567334) (20,-0.1511099453) (40,-0.1574670386)};
\addplot[color=decoder1,line width=1.5pt,dashed,mark=square*,mark size=1.4pt,mark options={fill=white,solid}] coordinates {(5,-0.0728301979) (10,-0.0719540804) (20,-0.0808072411) (40,-0.0915734522)};
\addplot[color=decoder2,line width=1.5pt,mark=*,mark size=1.5pt,mark options={fill=white,solid}] coordinates {(5,-0.0490945499) (10,-0.0686086321) (20,-0.0789405773) (40,-0.0788720180)};
\addplot[color=decoder2,line width=1.5pt,dashed,mark=square*,mark size=1.4pt,mark options={fill=white,solid}] coordinates {(5,-0.0744297553) (10,-0.0866396684) (20,-0.0786200806) (40,-0.0757487875)};
\addplot[only marks,mark=diamond*,mark size=3pt,mark options={fill=decoder1,draw=black,line width=.5pt,xshift=6pt}] coordinates {(40,-0.1490250069) (40,-0.0862663063)};
\end{groupplot}\node[anchor=north,font=\fontsize{8}{9}\selectfont,align=center] at ([yshift=-10mm,xshift=9mm]group c1r1.south east) {\tikz[baseline=-.5ex]{\draw[decoder0,line width=1.5pt] (0,0) -- (1.8em,0);\path[draw=decoder0,fill=white,line width=1pt] (.9em,0) circle (1.5pt);} Logistic\quad\tikz[baseline=-.5ex]{\draw[decoder1,line width=1.5pt] (0,0) -- (1.8em,0);\path[draw=decoder1,fill=white,line width=1pt] (.9em,0) circle (1.5pt);} CatBoost\quad\tikz[baseline=-.5ex]{\draw[decoder2,line width=1.5pt] (0,0) -- (1.8em,0);\path[draw=decoder2,fill=white,line width=1pt] (.9em,0) circle (1.5pt);} TabPFN-3\qquad\tikz[baseline=-.5ex]{\path[fill=decoder1,draw=black,line width=.5pt] (0,3pt) -- (2.25pt,0) -- (0,-3pt) -- (-2.25pt,0) -- cycle;} Context-tuned CatBoost (40 rows)};
\end{tikzpicture}%
\endgroup}
\caption{\textbf{GraphLog encoder effects across context budgets.}
(\textit{a}) Equal-world macro balanced accuracy for the relation-multiset (solid) and
absolute-position (dashed) encoders. (\textit{b})
Relation-multiset-minus-absolute-position (solid) and block-permuted-minus-aligned (dashed)
contrasts. The
diamonds mark context-tuned \textsc{CatBoost} at 40 rows for the two contrasts, drawn slightly to the right of the 40-row position.
Context rows per relation are shown on a logarithmic axis.}
\label{fig:graphlog_context}
\end{figure}

The 57 worlds draw subsets from 76 rules. Their mean pairwise rule-set Jaccard similarity is 0.230 and the maximum is 0.905. Similarity-cluster sensitivities preserve the direction of the logistic and nonlinear-decoder effects. Leave-cluster-out regressions based on rule overlap and simple world descriptors yield negative \(R^2\), indicating poor prediction of the observed encoder contrast from these descriptors. These checks describe robustness within the complete benchmark, with worlds treated as fixed benchmark units.

\begin{table}[!ht]
\centering
\scriptsize
\caption{\textbf{GraphLog row-equivalence diagnostics.}
Ordinary-accuracy ceilings and singleton mass are averaged equally across
worlds at each OOD length. The main decoder results use macro balanced accuracy and the corresponding class-reweighted ceiling.}
\label{tab:graphlog_ceiling_detail}
\begin{adjustbox}{max width=\linewidth}
\begin{tabular}{lrrrr}
\toprule
Encoder & Ceiling L4 & Singleton L4 & Ceiling L5 & Singleton L5 \\
\midrule
Length only & 0.290 & 0.000 & 0.269 & 0.000 \\
Relation multiset & 0.963 & 0.005 & 0.974 & 0.163 \\
Directed transitions & 0.981 & 0.007 & 0.990 & 0.188 \\
Absolute position & 0.981 & 0.007 & 0.992 & 0.188 \\
Absolute + multiset & 0.981 & 0.007 & 0.992 & 0.188 \\
\bottomrule
\end{tabular}
\end{adjustbox}
\end{table}

The class-reweighted ceilings used for the main GraphLog comparison are \GraphLogSharedCeiling{} for the relation multiset and \GraphLogAbsoluteCeiling{} for absolute positions. \cref{tab:graphlog_capacity} reports the capacity sweep. Every depth is rerun on all 57 worlds with the same context rows, evaluation queries, 500 trees, learning rate 0.05, and three seeds. Increasing depth raises context fit for both encoders while progressively reducing multiset evaluation accuracy. This pattern supports capacity-amplified overfitting of the multiset row under this \CatBoost{} configuration.

\begin{table}[!ht]
\centering
\scriptsize
\caption{\textbf{GraphLog \CatBoost{} capacity sweep.} Evaluation values are equal-world means of equal-length macro balanced accuracy over lengths 4--5. Context values are equal-world macro balanced accuracy. The effect is relation multiset minus absolute position.}
\label{tab:graphlog_capacity}
\begin{adjustbox}{max width=\linewidth}
\begin{tabular}{rrrrrrr}
\toprule
Depth & Eval. multiset & Eval. absolute & Effect & Context multiset & Context absolute & Multiset at/below chance \\
\midrule
1 & 0.201 & 0.271 & -0.070 & 0.830 & 0.910 & 5 \\
2 & 0.169 & 0.264 & -0.096 & 0.930 & 0.980 & 9 \\
3 & 0.150 & 0.263 & -0.112 & 0.945 & 0.988 & 20 \\
4 & 0.138 & 0.268 & -0.129 & 0.947 & 0.990 & 25 \\
5 & 0.129 & 0.278 & -0.149 & 0.948 & 0.991 & 25 \\
6 & 0.126 & 0.282 & -0.156 & 0.948 & 0.991 & 27 \\
7 & 0.125 & 0.282 & -0.157 & 0.948 & 0.991 & 30 \\
\bottomrule
\end{tabular}
\end{adjustbox}
\end{table}

At 40 examples per class, multiset and absolute-position performance are at or below each world's chance level in 0 and 3 worlds for logistic regression, 30 and 1 for depth-7 \CatBoost{}, and 12 and 4 for \TabPFNthree{}.

\subsection{Dihedral Boundary Control}

The independent control samples ten generated populations over the 16 symmetries of a regular octagon. Context sequences have lengths 2--3, evaluation sequences have lengths 4--10, and the ordered group product is the target. The absolute-position encoder preserves the element at each sequence position. The coarse encoder removes order and merges elements separated by four rotations, so it discards both ordering information and part of the group-element identity.

Across generated populations, the 95\% sensitivity ranges for coarse-minus-absolute accuracy are \([-0.0028,0.0020]\), \([0.0106,0.0147]\), and \([-0.0042,0.0023]\) for logistic regression, \CatBoost{}, and \TabPFNthree{}. Both encoders remain at or near chance apart from the small \CatBoost{} contrast. The 257-column directed-transition encoder also remains at floor, reaching \DEightLogTransition{}/\DEightCatTransition{}/\DEightTabTransition{} for the three decoders. This boundary control limits the CLUTRR interpretation: when coarsening also removes task-relevant identity, both representations remain at floor apart from the small \CatBoost{} contrast.

\section{Knowledge Graph Panel}
\label{app:ultra57}

This section documents the panel, encoders, context constructions, and attribution checks underlying \cref{sec:feature-contributions} and \cref{sec:endpoint-connectivity}. The presentation separates the fixed evaluation protocol from the encoder and context variants so that every reported contrast can be traced to one changed component.

\subsection{Panel Composition and Evaluation Protocol}

The panel contains 57 knowledge graphs grouped into 16 metric-free provenance families. Family assignment follows source lineage and is independent of decoder outcomes. \cref{tab:ultra_membership} lists the complete panel.

\begin{table}[!ht]
\centering
\scriptsize
\caption{\textbf{Summary of the 57-KG panel.}
Dataset names reproduce the fixed asset identifiers. Provenance families define
the equal-weight aggregation and resampling units used in the main analysis.}
\label{tab:ultra_membership}
\begin{adjustbox}{max width=\linewidth}
\begin{tabular}{p{0.16\linewidth}p{0.78\linewidth}}
\toprule
Family & Dataset identifiers \\
\midrule
AristoV4 & \texttt{aristov4} \\
CoDEx & \texttt{codex\_large}, \texttt{codex\_medium}, \texttt{codex\_small} \\
ConceptNet100K & \texttt{conceptnet100k} \\
DBpedia100K & \texttt{dbpedia100k} \\
FB15k-237 & \texttt{fb15k237}, \texttt{fb15k237\_10}, \texttt{fb15k237\_20}, \texttt{fb15k237\_50}, \texttt{fb15k237\_ind\_v1}, \texttt{fb15k237\_ind\_v2}, \texttt{fb15k237\_ind\_v3}, \texttt{fb15k237\_ind\_v4}, \texttt{fb\_ingram\_25}, \texttt{fb\_ingram\_50}, \texttt{fb\_ingram\_75}, \texttt{fb\_ingram\_100} \\
FBNELL & \texttt{fbnell} \\
Hetionet & \texttt{hetionet} \\
H\&M & \texttt{hm\_1k}, \texttt{hm\_3k}, \texttt{hm\_5k}, \texttt{hm\_indigo} \\
ILPC2022 & \texttt{ilpc2022\_large}, \texttt{ilpc2022\_small} \\
MetaFam & \texttt{metafam} \\
NELL & \texttt{nell23k}, \texttt{nell995}, \texttt{nell\_ind\_v1}, \texttt{nell\_ind\_v2}, \texttt{nell\_ind\_v3}, \texttt{nell\_ind\_v4}, \texttt{nl\_ingram\_0}, \texttt{nl\_ingram\_25}, \texttt{nl\_ingram\_50}, \texttt{nl\_ingram\_75}, \texttt{nl\_ingram\_100} \\
WDSinger & \texttt{wdsinger} \\
Wiki-Topics & \texttt{wiki\_mt1\_health}, \texttt{wiki\_mt1\_tax}, \texttt{wiki\_mt2\_org}, \texttt{wiki\_mt2\_sci}, \texttt{wiki\_mt3\_art}, \texttt{wiki\_mt3\_infra}, \texttt{wiki\_mt4\_health}, \texttt{wiki\_mt4\_sci} \\
WK-InGram & \texttt{wk\_ingram\_25}, \texttt{wk\_ingram\_50}, \texttt{wk\_ingram\_75}, \texttt{wk\_ingram\_100} \\
WN18RR & \texttt{wn18rr}, \texttt{wn18rr\_ind\_v1}, \texttt{wn18rr\_ind\_v2}, \texttt{wn18rr\_ind\_v3}, \texttt{wn18rr\_ind\_v4} \\
YAGO3-10 & \texttt{yago310} \\
\bottomrule
\end{tabular}
\end{adjustbox}
\end{table}

For each dataset and seed, the source pool for the complete encoder ladder contains 400 positives sampled without replacement from the official held-out target triples and 400 endpoint corruptions that retain the positive relation. The budgets of 20, 50, 100, and 200 examples per binary class are deterministic nested subsets. No relation is dropped or duplicated to satisfy a per-relation quota. These baseline corruptions are not guaranteed to exclude every known positive triple and are distinct from the explicitly filtered condition in \cref{app:negative-context}.

The largest 400-row local context matches the inference-time local-context size of TFMLinker~\citep{liao2026tfmlinker}, which uses 200 positive and 200 negative local rows together with 80 learned global prototypes. Our setup retains the 400-row local context while freezing upstream models and auditing labelled target-graph context, leaving the learned cross-graph prototypes outside the comparison.

A lower-bound audit finds that 32 base negative rows coincide with one of the selected 64-query evaluation positives (0.047\% of the 68,400 base negative rows). After expansion over the four nested context budgets, 19 selected incidences occur across 8 of 684 dataset--seed--budget cells, corresponding to \(0.030\%\) of selected negative-row incidences. This audit covers overlap with the sampled evaluation positives.

Constructed features and frozen \Ultra{} scores use only the observed support graph, which excludes every held-out target triple. Neither the queried target edge nor an artificial inverse copy is inserted when constructing its row, while naturally observed reverse-relation edges remain available. Across all 684 dataset--seed--budget contexts, selected positive context triples and the 64 sampled evaluation triples have zero overlap. Evaluation samples 64 official test triples stratified by relation and ranks both head and tail prediction against all entities under standard filtered masks. Every paired comparison uses the same queries, candidates, filtering, and tie handling.

Let \(M_{d,s}\) denote the query-averaged metric for dataset \(d\) and seed \(s\), and let \(\mathcal D_f\) be the datasets assigned to provenance family \(f\). The equal-family estimate is
\begin{equation}
\overline M=\frac{1}{|\mathcal F|}\sum_{f\in\mathcal F}\frac{1}{|\mathcal D_f|}\sum_{d\in\mathcal D_f}\frac{1}{|\mathcal S|}\sum_{s\in\mathcal S}M_{d,s}.
\label{eq:equal-family-mean}
\end{equation}
\cref{eq:equal-family-mean} prevents large provenance families from dominating the panel average. Provenance families are also the resampling and sign-flip units.

Because eight of the sixteen families contain a single dataset, we report equal-dataset aggregation alongside the registered equal-family estimand. For the full-\TabPFNthree{} minus native-\Ultra{} contrast, the equal-family mean is \UltraFullNative{} and the equal-dataset mean is \UltraFullNativeDatasetMean{}. Their medians are \UltraFullNativeFamilyMedian{} and \UltraFullNativeDatasetMedian{}. Dataset wins/ties/losses are \UltraFullNativeWTL{} under the declared \(\pm0.005\) band. The leave-one-family-out range is \UltraFullNativeLOFO{}, and removing MetaFam yields \UltraFullNativeNoMeta{}. These are fixed-panel sensitivities.

\subsection{Query-Relative Pointwise Control}
\label{app:query-relative}

The tabular decoders are trained as pointwise binary classifiers on a shared labelled context but are evaluated by filtered ranking within each directional query. Their training loss targets pointwise classification, whereas MRR depends on within-query candidate ordering; classification and ranking losses therefore require distinct analyses~\citep{menon2016bipartite}. Cross-query calibration is irrelevant to MRR, which is computed separately within each query. We therefore repeat the 200-per-class comparison using two label-free, query-relative feature maps. For each numerical column, \emph{within-query z-score} subtracts the group mean and divides by its standard deviation, mapping constant columns to zero. \emph{Within-query percentile} replaces values by tie-aware midranks scaled to \([0,1]\). The maps are applied independently to every directional group: a context group contains a source positive and its paired endpoint corruption, while an evaluation group contains all entity candidates for one head- or tail-prediction query. Thus context labels and evaluation targets are never used by the transformation.

The control crosses the raw, z-score, and percentile maps with the \emph{Local}, \emph{score}, and \emph{Local + score} encoders and all three decoders, using the same 57 datasets, three seeds, pool-random positive contexts with filtered-uniform negatives, queries, candidates, and ranking masks. These positive contexts differ from the evaluation-relation-matched positive contexts used in \cref{tab:context_matrix}. Because an affine or rank transform of the one-column frozen score is order preserving within each query, score-only logistic regression must reproduce native \Ultra{} whenever the fitted coefficient has positive orientation. Exact native-rank equality is checked cell by cell. The remaining contrasts test whether query-relative inputs remedy the mismatch for a nonlinear or multivariate pointwise readout.

\cref{tab:query-relative-control} reports all \QueryRelCells{} completed cells. Within-query scaling improves the \emph{Local} row for logistic regression, but not the multivariate \emph{Local + score} row. It lowers MRR for all six \CatBoost{} transformations and all six \TabPFNthree{} transformations. All six \CatBoost{} reductions and five of the six \TabPFNthree{} reductions remain supported after joint correction. The remaining \TabPFNthree{} percentile contrast is negative but does not survive the correction. For score-only logistic regression, all \QueryRelRankCells{} dataset--seed cells reproduce native \Ultra{} exactly under every map, covering \QueryRelRanks{} directional query rankings per map. These controls show that the effect of query-relative scaling is decoder-specific. The analysis is scoped to the tested pointwise readouts under the fixed ranking protocol.

\begin{table}[!ht]
\centering
\scriptsize
\caption{\textbf{Query-relative control.} Entries are equal-family MRR. Parentheses report the transformed-minus-raw effect. Z-score and percentile maps are applied by the same label-free rule to context groups and evaluation candidate blocks.}
\label{tab:query-relative-control}
\setlength{\tabcolsep}{4.5pt}
\begin{adjustbox}{max width=\linewidth}
\begin{tabular}{llrrr}
\toprule
Decoder & Encoder & Raw & Within-query z-score & Within-query percentile \\
\midrule
\multirow{3}{*}{Logistic}
 & Local & \QueryRelLogLocalRaw{} & \QueryRelLogLocalZ{} (\QueryRelLogLocalZDelta{}) & \QueryRelLogLocalPct{} (\QueryRelLogLocalPctDelta{}) \\
 & Score & \QueryRelLogScoreRaw{} & \QueryRelLogScoreZ{} (0.000) & \QueryRelLogScorePct{} (0.000) \\
 & Local + score & \QueryRelLogLocalScoreRaw{} & \QueryRelLogLocalScoreZ{} (\QueryRelLogLocalScoreZDelta{}) & \QueryRelLogLocalScorePct{} (\QueryRelLogLocalScorePctDelta{}) \\
\midrule
\multirow{3}{*}{\CatBoost{}}
 & Local & \QueryRelCatLocalRaw{} & \QueryRelCatLocalZ{} (\QueryRelCatLocalZDelta{}) & \QueryRelCatLocalPct{} (\QueryRelCatLocalPctDelta{}) \\
 & Score & \QueryRelCatScoreRaw{} & \QueryRelCatScoreZ{} (\QueryRelCatScoreZDelta{}) & \QueryRelCatScorePct{} (\QueryRelCatScorePctDelta{}) \\
 & Local + score & \QueryRelCatLocalScoreRaw{} & \QueryRelCatLocalScoreZ{} (\QueryRelCatLocalScoreZDelta{}) & \QueryRelCatLocalScorePct{} (\QueryRelCatLocalScorePctDelta{}) \\
\midrule
\multirow{3}{*}{\TabPFNthree{}}
 & Local & \QueryRelTabLocalRaw{} & \QueryRelTabLocalZ{} (\QueryRelTabLocalZDelta{}) & \QueryRelTabLocalPct{} (\QueryRelTabLocalPctDelta{}) \\
 & Score & \QueryRelTabScoreRaw{} & \QueryRelTabScoreZ{} (\QueryRelTabScoreZDelta{}) & \QueryRelTabScorePct{} (\QueryRelTabScorePctDelta{}) \\
 & Local + score & \QueryRelTabLocalScoreRaw{} & \QueryRelTabLocalScoreZ{} (\QueryRelTabLocalScoreZDelta{}) & \QueryRelTabLocalScorePct{} (\QueryRelTabLocalScorePctDelta{}) \\
\bottomrule
\end{tabular}
\end{adjustbox}
\end{table}

\subsection{Positive-Context Relation Coverage}
\label{app:relation-context}

We additionally vary the positive context while keeping its size and the evaluation queries fixed. \emph{Pool-random positives} are sampled from the available source pool, whereas \emph{evaluation-relation-matched positives} are selected to better match the relation distribution of the evaluation queries. Relation matching increases equal-family relation-type coverage from \(0.667\) to \(0.871\) and relation-mass coverage from \(0.693\) to \(0.891\), while reducing the total-variation distance between context and evaluation relation distributions from \(0.447\) to \(0.190\).

Despite these large changes in relational coverage, the corresponding MRR effects are generally modest across the tested encoder and decoder configurations. Among the 12 evaluation-relation-matched-minus-pool-random contrasts, only the \CatBoost{} local encoder without relation identity remains supported after the joint multiple-comparison correction, with an equal-family MRR change of \(+0.0139\) (\(p_{\max T}=0.0369\)). Increasing relation coverage therefore changes the labelled support substantially, while the corresponding MRR effects remain modest.

\subsection{Relation-Identity Ablation}
\label{app:relation-id}

Because the \emph{Local} encoder includes relation identity whereas the entity-token control does not, we separately remove the relation-identity column and repeat the matched comparisons across decoders, upstream-score conditions, and positive-context constructions. Across the relation-identity contrasts, the absolute equal-family MRR effect
never exceeds $0.0031$, and none remains supported after the joint max-$|t|$
correction. The relation-identity column therefore accounts for little of the observed local-versus-identifier difference under these matched conditions.

\subsection{Encoder Schemas}

\cref{tab:interfaces} defines the nine encoders used in the complete encoder ladder and in the matched additions reported in \cref{sec:feature-contributions}.

\begin{table}[!ht]
\centering
\small
\caption{\textbf{Graph-to-table encoder schemas.}
The nine encoders are grouped by the evidence exposed by their columns.
Composite encoders concatenate the named blocks without recomputation and
inherit the icons of their constituent evidence families.}
\label{tab:interfaces}
\begin{adjustbox}{max width=\linewidth}
\begin{threeparttable}
\begin{tabular}{@{}llrl@{}}
\toprule
& \textbf{Encoder} & \textbf{Dim.} & \textbf{Columns or blocks} \\
\midrule
\evOpaque & Entity tokens & 2 & Dataset-scoped head and tail tokens \\
\addlinespace[2pt]
\evLocal & Dyadic & 3 & Other-relation pair count, reverse-pair count, same entity \\
\evLocal & Paths & 4 & Two-hop, start-relation, end-relation, reverse-two-hop counts \\
\evLocal & Local & 18 & Relation identity; endpoint, relation, pair, path, community statistics \\
\addlinespace[2pt]
\evGlobal & Global & 13 & PageRank, component, heat, diffusion \\
\addlinespace[2pt]
\evScore & \Ultra{} score & 1 & Frozen \Ultra{}-4G candidate score \\
\addlinespace[3pt]
\hdashline
\addlinespace[3pt]
\evLocalGlobal & Local + global & 31 & 18 local \(\oplus\) 13 global \\
\evLocalScore & Local + score & 19 & 18 local \(\oplus\) frozen score \\
\evFull & Full table & 32 & 18 local \(\oplus\) frozen score \(\oplus\) 13 global \\
\bottomrule
\end{tabular}
\begin{tablenotes}
\small
\item[~]\evOpaque = opaque identifiers; \evLocal = local relational structure; \evGlobal = global topological and spectral structure; \evScore = learned-model score.
\end{tablenotes}
\end{threeparttable}
\end{adjustbox}
\end{table}

The 18 local columns are relation identity; total in- and out-degree for each endpoint; head relation-out and tail relation-in degree; relation frequency, functionality, and inverse functionality; other-relation pair and reverse-pair counts; two-hop, start-relation, end-relation, and reverse-two-hop counts; shared community assignment; and community relation density. The 13 global columns are head and tail PageRank, membership in the same weakly connected component, head and tail log component fraction, and truncated low-frequency heat-projector similarity and diffusion distance at \(t\in\{0.1,1,10,100\}\). The community and weak-component indicators are distinct, and no column is duplicated.

\subsection{Complete Encoder Ladder}

\cref{tab:ultra_ladder_complete} reports the absolute \TabPFNthree{} levels behind the nested comparisons in \cref{sec:feature-contributions}. Dyadic evidence captures much of the narrow endpoint-pair signal, paths improve with context but remain weaker in aggregate, global features alone remain limited, and the frozen score is the strongest individual feature group. Composite levels combine these sources and are therefore interpreted through the paired additions in \cref{app:paired-context-effects}.

\begin{table}[!ht]
\centering
\scriptsize
\caption{\textbf{\textsc{TabPFN-3} performance across the encoder
ladder.} Entries report equal-family MRR under filtered candidate evaluation
on the 57-KG panel. The \emph{unfiltered-uniform} context negatives are
target-excluding endpoint corruptions and are not filtered by the
all-known-positive mask. Contexts
contain 20, 50, 100, or 200 examples per binary class.}
\label{tab:ultra_ladder_complete}
\begin{adjustbox}{max width=\linewidth}
\begin{tabular}{lrrrr}
\toprule
Encoder & 20 & 50 & 100 & 200 \\
\midrule
Dataset-scoped entity IDs & 0.002 & 0.002 & 0.003 & 0.004 \\
Dyadic & 0.119 & 0.136 & 0.149 & 0.152 \\
Paths & 0.064 & 0.068 & 0.071 & 0.076 \\
Local & 0.167 & 0.208 & 0.228 & 0.237 \\
Global & 0.053 & 0.059 & 0.069 & 0.068 \\
\Ultra{} score & 0.299 & 0.305 & 0.307 & 0.301 \\
Local + global & 0.134 & 0.189 & 0.217 & 0.233 \\
Local + score & 0.297 & 0.323 & 0.334 & 0.336 \\
Full & 0.287 & 0.321 & 0.339 & 0.341 \\
\bottomrule
\end{tabular}
\end{adjustbox}
\end{table}

\subsection{Negative-Context Construction and Audit Checks}
\label{app:negative-context}

The context-robustness experiment fixes 200 positive and 200 negative context rows for each dataset and seed. Context positives and the 64 sampled evaluation triples are disjoint. Positive rows and evaluation queries are identical across the two negative-context conditions.

Filtered-uniform negatives replace the missing endpoint with a uniformly sampled candidate excluded by the all-known-positive mask. Structurally plausible negatives first restrict candidates to the appropriate relation-specific endpoint role and, when possible, the same graph partition as the fixed endpoint. They then sample among the 32 closest candidates under relative in- and out-degree. Empty role and partition pools trigger explicit fallbacks, observed for 1,399 and 14,446 selected negatives (4.1\% and 42.2\% of the 34,200). The selector uses only the graph-based criteria above. Its selected negatives nevertheless receive a mean frozen score of \(-1.238\), compared with \(-6.929\) under filtered-uniform sampling, indicating that the two conditions differ in upstream-model difficulty despite sharing positives and evaluation.

All 171 dataset--seed context audits contain zero known-positive negative rows, and all positive-row hashes agree between conditions. Evaluation indices, candidate sets, and filtered masks are unchanged. The paired contrast therefore isolates the negative context distribution.

The complete encoder ladder in \cref{tab:ultra_ladder_complete} uses endpoint corruptions without an all-known-positive rejection step, whereas the context-robustness experiment uses filtered negatives. We isolate filtering with a separate matched asset constructed from one deterministic corruption stream. Across its unfiltered and filtered arms, positive rows, directional query identities, seeds, evaluation candidates, masks, targets, and feature schemas are identical. Only rejection of candidates in the all-known-positive set changes. The unfiltered arm contains 456 known-positive negatives among 34,200 rows (1.33\%), distributed across 88 of 171 dataset--seed contexts. Contamination reaches 39.5\% for one seed of \texttt{nell\_ind\_v1}. We rerun \TabPFNthree{} on the \emph{Local}, \emph{score}, and \emph{Local + score} encoders in all \FilterCells{} matched cells.

\begin{table}[!ht]
\centering
\scriptsize
\caption{\textbf{Matched all-known-positive filtering control.} The effect is filtered-uniform minus unfiltered-uniform at 200 examples per class. Levels and effects are MRR; sensitivities resample the 16 provenance families. The last column is the exact family-level max-$|t|$ correction across the three contrasts.}
\label{tab:matched-filter-control}
\setlength{\tabcolsep}{4pt}
\begin{adjustbox}{max width=\linewidth}
\begin{tabular}{lrrrrrr}
\toprule
Encoder & Unfiltered & Filtered & Equal-family & Equal-dataset & 95\% sensitivity & $p_{\max T}$ \\
\midrule
Local & \FilterLocalUnfiltered{} & \FilterLocalFiltered{} & \FilterLocalFamilyDelta{} & \FilterLocalDatasetDelta{} & \FilterLocalSensitivity{} & \FilterLocalMaxP{} \\
Score & \FilterScoreUnfiltered{} & \FilterScoreFiltered{} & \FilterScoreFamilyDelta{} & \FilterScoreDatasetDelta{} & \FilterScoreSensitivity{} & \FilterScoreMaxP{} \\
Local + score & \FilterLocalScoreUnfiltered{} & \FilterLocalScoreFiltered{} & \FilterLocalScoreFamilyDelta{} & \FilterLocalScoreDatasetDelta{} & \FilterLocalScoreSensitivity{} & \FilterLocalScoreMaxP{} \\
\bottomrule
\end{tabular}
\end{adjustbox}
\end{table}

Filtering yields a modest \FilterLocalFamilyDelta{} equal-family improvement for \emph{Local}, with sensitivity \FilterLocalSensitivity{}, but it falls just outside the jointly corrected threshold ($p_{\max T}=\FilterLocalMaxP{}$). The score-containing effects are smaller and unsupported. The contamination audit is therefore operationally important, especially for NELL-ind-v1, while the matched filtering effects are substantially smaller than the cross-table differences of 0.030--0.055 MRR.

\subsection{Paired Context and Encoder Effects}
\label{app:paired-context-effects}

\cref{tab:context_matrix} gives the absolute equal-family levels associated with \cref{fig:context_robustness}. The best decoder depends on the encoder. Logistic regression is strongest for the score-only row, while \TabPFNthree{} is strongest for local and local-plus-score rows under both context constructions.

\begin{table}[!ht]
\centering
\scriptsize
\caption{\textbf{Context-robustness results.}
Entries report equal-family MRR at 200 examples per class for filtered-uniform and structurally plausible negative contexts.
Evaluation-relation-matched positive rows and evaluation queries are identical across conditions.
Bold indicates the best decoder within each encoder and context.}
\label{tab:context_matrix}
\setlength{\tabcolsep}{4.5pt}
\begin{adjustbox}{max width=\linewidth}
\begin{tabular}{lrrr@{\hspace{10pt}}rrr}
\toprule
& \multicolumn{3}{c}{Filtered uniform} & \multicolumn{3}{c}{Structurally plausible} \\
\cmidrule(lr){2-4}\cmidrule(lr){5-7}
Encoder & Logistic & \CatBoost{} & \TabPFNthree{} & Logistic & \CatBoost{} & \TabPFNthree{} \\
\midrule
Local & \ContextLogLocalUniform{} & \ContextCatLocalUniform{} & \textbf{\ContextTabLocalUniform{}} & \ContextLogLocalStructural{} & \ContextCatLocalStructural{} & \textbf{\ContextTabLocalStructural{}} \\
Score & \textbf{\ContextLogScoreUniform{}} & \ContextCatScoreUniform{} & \ContextTabScoreUniform{} & \textbf{\ContextLogScoreStructural{}} & \ContextCatScoreStructural{} & \ContextTabScoreStructural{} \\
Local + score & \ContextLogLocalScoreUniformLevel{} & \ContextCatLocalScoreUniformLevel{} & \textbf{\ContextTabLocalScoreUniformLevel{}} & \ContextLogLocalScoreStructuralLevel{} & \ContextCatLocalScoreStructuralLevel{} & \textbf{\ContextTabLocalScoreStructuralLevel{}} \\
\bottomrule
\end{tabular}
\end{adjustbox}
\end{table}

For each contrast \(c\), let \(\delta_{f,c}\) be its mean paired effect within provenance family \(f\). The exact multiple-comparison procedure enumerates all \(2^{16}\) family-level sign assignments, recomputes the one-sample \(t\)-statistic for every contrast, and records \(T=\max_c|t_c|\). The adjusted \(p_{\mathrm{max}T}\) in \cref{tab:context_effects_full} is the fraction of sign assignments whose maximum statistic is at least as large as the observed \(|t_c|\). This preserves the family as the inferential unit and controls the 30 tested panel contrasts jointly.

\begin{table}[!ht]
\centering
\scriptsize
\caption{\textbf{Paired encoder effects.}
Effects are equal-family $\Delta\mathrm{MRR}$ with 95\% provenance-family
bootstrap sensitivity intervals. Positive values favour the first
term in each contrast. Dataset wins, ties, and losses use a $\pm 0.005$ band.
The final column reports the exact max-$|t|$ family-level sign-flip correction
across 30 contrasts.}
\label{tab:context_effects_full}
\setlength{\tabcolsep}{4pt}
\begin{adjustbox}{max width=\linewidth}
\begin{tabular}{lllrrrr}
\toprule
Contrast & Context & Decoder & Effect & 95\% sensitivity & W/T/L & \(p_{\mathrm{max}T}\) \\
\midrule
\multirow{6}{*}{Local minus entity IDs}
 & Filtered uniform & Logistic & +0.197 & [+0.117,+0.314] & 57/0/0 & 0.018 \\
 & Filtered uniform & \CatBoost{} & +0.258 & [+0.177,+0.372] & 57/0/0 & 0.002 \\
 & Filtered uniform & \TabPFNthree{} & +0.262 & [+0.184,+0.378] & 57/0/0 & 0.001 \\
 & Structural & Logistic & +0.177 & [+0.097,+0.301] & 57/0/0 & 0.045 \\
 & Structural & \CatBoost{} & +0.194 & [+0.118,+0.314] & 57/0/0 & 0.019 \\
 & Structural & \TabPFNthree{} & +0.208 & [+0.128,+0.329] & 57/0/0 & 0.013 \\
\midrule
\multirow{6}{*}{Score added to local}
 & Filtered uniform & Logistic & +0.129 & [+0.093,+0.162] & 56/0/1 & <0.001 \\
 & Filtered uniform & \CatBoost{} & +0.048 & [+0.033,+0.065] & 48/3/6 & <0.001 \\
 & Filtered uniform & \TabPFNthree{} & +0.124 & [+0.096,+0.152] & 54/3/0 & <0.001 \\
 & Structural & Logistic & +0.083 & [+0.041,+0.129] & 41/0/16 & 0.017 \\
 & Structural & \CatBoost{} & +0.061 & [+0.030,+0.096] & 36/5/16 & 0.018 \\
 & Structural & \TabPFNthree{} & +0.056 & [+0.023,+0.092] & 38/3/16 & 0.051 \\
\midrule
\multirow{6}{*}{Local added to score}
 & Filtered uniform & Logistic & -0.007 & [-0.065,+0.095] & 4/6/47 & 1.000 \\
 & Filtered uniform & \CatBoost{} & +0.012 & [-0.054,+0.125] & 10/2/45 & 1.000 \\
 & Filtered uniform & \TabPFNthree{} & +0.059 & [+0.005,+0.158] & 28/12/17 & 0.978 \\
 & Structural & Logistic & -0.039 & [-0.115,+0.076] & 10/3/44 & 1.000 \\
 & Structural & \CatBoost{} & +0.064 & [-0.003,+0.168] & 39/3/15 & 0.955 \\
 & Structural & \TabPFNthree{} & +0.024 & [-0.038,+0.128] & 18/6/33 & 1.000 \\
\bottomrule
\end{tabular}
\end{adjustbox}
\end{table}

\cref{tab:context_effects_full} formalises the asymmetry discussed in \cref{sec:feature-contributions}. Adding the frozen score to local features is positive in all six conditions and remains supported in five after joint correction. Adding local features to the score is smaller and unstable, with no adjusted contrast supported. The direction of a feature addition must therefore be retained when interpreting nested encoders.

\begin{table}[!ht]
\centering
\scriptsize
\caption{\textbf{Effect of negative-context construction.}
Filtered-uniform negatives are replaced with structurally plausible negatives
while positive rows and evaluation queries remain fixed. Entries report
equal-family $\Delta\mathrm{MRR}$ with 95\% provenance-family bootstrap
sensitivity intervals and dataset wins, ties, and losses under a $\pm 0.005$ band.
Positive values favour the structurally plausible context.}
\label{tab:context_shift_full}
\setlength{\tabcolsep}{4pt}
\begin{adjustbox}{max width=\linewidth}
\begin{tabular}{llrrr}
\toprule
Encoder & Decoder & Effect & 95\% sensitivity & W/T/L \\
\midrule
Local & Logistic & -0.020 & [-0.032,-0.010] & 10/8/39 \\
 & \CatBoost{} & -0.063 & [-0.084,-0.042] & 4/2/51 \\
 & \TabPFNthree{} & -0.054 & [-0.075,-0.033] & 6/3/48 \\
\midrule
Score & Logistic & -0.033 & [-0.073,-0.005] & 0/45/12 \\
 & \CatBoost{} & -0.102 & [-0.144,-0.060] & 8/1/48 \\
 & \TabPFNthree{} & -0.086 & [-0.132,-0.043] & 4/13/40 \\
\midrule
Local + score & Logistic & -0.066 & [-0.102,-0.032] & 10/4/43 \\
 & \CatBoost{} & -0.050 & [-0.087,-0.014] & 13/1/43 \\
 & \TabPFNthree{} & -0.122 & [-0.164,-0.081] & 1/2/54 \\
\bottomrule
\end{tabular}
\end{adjustbox}
\end{table}

\cref{tab:context_shift_full} shows that structurally plausible negatives lower MRR for every informative encoder--decoder combination. The identifier baselines change by less than 0.001 and are omitted from the table. Since positives and evaluation remain fixed, these effects quantify sensitivity to the negative-context construction \(\gttS\).

\subsection{Upstream-Score Lineage and Native-Ranking Checks}

The frozen \texttt{ultra\_4g} checkpoint was pretrained on FB15k-237, WN18RR, CoDEx-Medium, and NELL-995. These lineages occur in the 57-KG panel. Comparisons involving the frozen score therefore audit a fixed upstream model on a panel that includes some of its pretraining lineages.

As a direct-lineage sensitivity, we remove the CoDEx, FB15k-237, FBNELL, NELL, and WN18RR provenance families. The local-versus-identifier and score-given-local effects remain positive in every retained family. This sensitivity specifically addresses known dataset lineage.

For score-only logistic regression, every fitted coefficient has positive orientation. Ranking the original \Ultra{} score directly consequently preserves its ordering and ties by construction. The complete rerun verifies exact equality between logistic and native ranks across all \UltraScoreRankAuditCells{} combinations of dataset, context budget, and seed, with maximum rank difference zero.

\begin{table}[!ht]
\centering
\scriptsize
\caption{\textbf{Attribution and sensitivity checks for the frozen ULTRA
score.} Effects report equal-family MRR differences. Dataset and family counts
indicate how often \textsc{TabPFN-3} with the \emph{Full} encoder performs below
native ULTRA ranking. A dash denotes a quantity not used in the corresponding
comparison.}
\label{tab:native_score_checks}
\begin{adjustbox}{max width=\linewidth}
\begin{tabular}{lrrrr}
\toprule
Comparison & Effect & 95\% sensitivity & Datasets below native & Families below native \\
\midrule
Global added after local + score & \UltraGlobalGivenScore{} & \UltraGlobalGivenScoreCI{} & -- & -- \\
Full \TabPFNthree{} minus native \Ultra{} & \UltraFullNative{} & \UltraFullNativeCI{} & 48/57 & 14/16 \\
Full minus native, excluding MetaFam & \UltraFullNativeNoMeta{} & -- & -- & -- \\
\bottomrule
\end{tabular}
\end{adjustbox}
\end{table}

\cref{tab:native_score_checks} provides the attribution checks summarised in \cref{sec:feature-contributions}. Global PageRank, component, heat, and diffusion features add only \UltraGlobalGivenScore{} once the score is present and do not remain supported after correction. \TabPFNthree{} with the \emph{Full} encoder underperforms native ranking on 48 of 57 datasets and 14 of 16 families. Its positive aggregate contrast is driven by MetaFam and becomes \UltraFullNativeNoMeta{} when that family is excluded.

A paired sensitivity analysis using 512 sampled test triples per dataset preserves the sign of all five panel-level encoder contrasts used in the paper. Mean changes relative to the 64-triple evaluation range from \(-0.0015\) to \(+0.0064\) MRR. The 64-triple sample is therefore used for paired panel conclusions.

\subsection{Query-Evidence Strata}
\label{app:query-evidence}

\cref{sec:endpoint-connectivity} groups directional evaluation queries by endpoint connectivity in the observed support graph. For query \(q=(h,r,t)\), let \(a_{h\rightarrow t}=1\) when the support graph contains at least one observed edge \((h,r',t)\) with \(r'\neq r\), and let \(a_{t\rightarrow h}\) be defined analogously for \((t,r',h)\). The held-out target edge is removed before these indicators are computed.

\begin{table}[!ht]
\centering
\scriptsize
\caption{\textbf{Endpoint-connectivity strata.}
$a_{h\to t}$ and $a_{t\to h}$ indicate whether another observed relation
connects the endpoints in the corresponding direction after removing the
held-out target edge. Counts cover the 7,296 directional queries obtained from
head and tail prediction for 64 test triples in each of 57 knowledge graphs.
Percentages are computed over the fixed panel.}
\label{tab:query_evidence_groups}
\begin{adjustbox}{max width=\linewidth}
\begin{tabular}{lccrr}
\toprule
Group & \(a_{h\rightarrow t}\) & \(a_{t\rightarrow h}\) & Queries & Share \\
\midrule
Both directions & 1 & 1 & \QueryBoth{} & 3.3\% \\
Reverse only & 0 & 1 & 964 & 13.2\% \\
Direct only & 1 & 0 & \QueryDirect{} & 5.7\% \\
No pair edge & 0 & 0 & \QueryNeither{} & 77.8\% \\
\bottomrule
\end{tabular}
\end{adjustbox}
\end{table}

\begin{figure}[!ht]
\centering
\resizebox{\linewidth}{!}{\begingroup
\fontfamily{ptm}\selectfont
\definecolor{plotbackground}{HTML}{FAF5F3}%
\colorlet{decoder0}{blue!80!magenta!60!white}%
\definecolor{decoder1}{HTML}{E0CB68}%
\colorlet{decoder2}{red!75!orange}%
\definecolor{encoderblue}{HTML}{1F65D6}%
\colorlet{enc-dyadic}{encoderblue!22!white}%
\colorlet{enc-paths}{encoderblue!58!white}%
\colorlet{enc-local}{encoderblue}%
\begin{tikzpicture}
\begin{groupplot}[group style={group size=3 by 1,horizontal sep=4mm},
scale only axis,width=3.55cm,height=4.6cm,xmin=0,xmax=.85,ymin=-.55,ymax=3.55,
xtick={0,.2,.4,.6,.8},ytick={3,2,1,0},yticklabels={},xmajorgrids=true,ymajorgrids=false,
grid=both,xlabel={Equal-family MRR},
grid style=dashed,axis background/.style={fill=plotbackground},
    every tick label/.append style={font=\fontsize{8}{8}\selectfont},
    scaled ticks=false,tick label style={/pgf/number format/fixed,/pgf/number format/precision=2},
    xlabel style={font=\small},ylabel style={font=\small},title style={font=\small,yshift=2pt},label style={font=\fontsize{8}{8}\selectfont}]
\nextgroupplot[title={(a) Logistic},yticklabels={Both directions,Reverse only,Direct only,No pair edge}]
\addplot[xbar,bar shift=0pt,bar width=5pt,fill=enc-dyadic,draw=black,line width=.35pt,mark=none,error bars/.cd,x dir=both,x explicit,error bar style={draw=black!80,line width=.65pt},error mark=-,error mark options={rotate=90,draw=black!80,mark size=1.8pt,line width=.65pt}] coordinates {(0.6677932670,3.25) += (0.1198477142,0) -= (0.1252934344,0) (0.4577783669,2.25) += (0.1023487663,0) -= (0.1016082668,0) (0.2970858004,1.25) += (0.0967315398,0) -= (0.0953644951,0) (0.0017736270,0.25) += (0.0017085446,0) -= (0.0011188005,0)};
\addplot[xbar,bar shift=0pt,bar width=5pt,fill=enc-paths,draw=black,line width=.35pt,mark=none,error bars/.cd,x dir=both,x explicit,error bar style={draw=black!80,line width=.65pt},error mark=-,error mark options={rotate=90,draw=black!80,mark size=1.8pt,line width=.65pt}] coordinates {(0.0685671602,3.00) += (0.0330467202,0) -= (0.0316474749,0) (0.0705830346,2.00) += (0.0496005993,0) -= (0.0352233213,0) (0.0458921038,1.00) += (0.0321281278,0) -= (0.0250779449,0) (0.0471354409,0.00) += (0.0146410263,0) -= (0.0132873886,0)};
\addplot[xbar,bar shift=0pt,bar width=5pt,fill=enc-local,draw=black,line width=.35pt,mark=none,error bars/.cd,x dir=both,x explicit,error bar style={draw=black!80,line width=.65pt},error mark=-,error mark options={rotate=90,draw=black!80,mark size=1.8pt,line width=.65pt}] coordinates {(0.5162283422,2.75) += (0.1090308323,0) -= (0.1081466765,0) (0.3778596138,1.75) += (0.1123103791,0) -= (0.1046480177,0) (0.2857261299,0.75) += (0.1166234672,0) -= (0.1107840379,0) (0.0428477939,-0.25) += (0.0124288101,0) -= (0.0116031265,0)};
\nextgroupplot[title={(b) CatBoost}]
\addplot[xbar,bar shift=0pt,bar width=5pt,fill=enc-dyadic,draw=black,line width=.35pt,mark=none,error bars/.cd,x dir=both,x explicit,error bar style={draw=black!80,line width=.65pt},error mark=-,error mark options={rotate=90,draw=black!80,mark size=1.8pt,line width=.65pt}] coordinates {(0.3819339523,3.25) += (0.1261292584,0) -= (0.1294407186,0) (0.4466936240,2.25) += (0.1338522373,0) -= (0.1399261054,0) (0.2942356979,1.25) += (0.0936071558,0) -= (0.0915346520,0) (0.0013846215,0.25) += (0.0009331586,0) -= (0.0007297960,0)};
\addplot[xbar,bar shift=0pt,bar width=5pt,fill=enc-paths,draw=black,line width=.35pt,mark=none,error bars/.cd,x dir=both,x explicit,error bar style={draw=black!80,line width=.65pt},error mark=-,error mark options={rotate=90,draw=black!80,mark size=1.8pt,line width=.65pt}] coordinates {(0.0454497020,3.00) += (0.0357526269,0) -= (0.0292611391,0) (0.0658502620,2.00) += (0.0414878834,0) -= (0.0306711104,0) (0.0534706094,1.00) += (0.0430022555,0) -= (0.0294030772,0) (0.0450368630,0.00) += (0.0135774225,0) -= (0.0119991485,0)};
\addplot[xbar,bar shift=0pt,bar width=5pt,fill=enc-local,draw=black,line width=.35pt,mark=none,error bars/.cd,x dir=both,x explicit,error bar style={draw=black!80,line width=.65pt},error mark=-,error mark options={rotate=90,draw=black!80,mark size=1.8pt,line width=.65pt}] coordinates {(0.4446312138,2.75) += (0.0946342884,0) -= (0.1054822131,0) (0.4105212420,1.75) += (0.1291853061,0) -= (0.1248915774,0) (0.2665636218,0.75) += (0.0868211637,0) -= (0.0894622713,0) (0.0823766746,-0.25) += (0.0171362532,0) -= (0.0156593445,0)};
\nextgroupplot[title={(c) TabPFN-3}]
\addplot[xbar,bar shift=0pt,bar width=5pt,fill=enc-dyadic,draw=black,line width=.35pt,mark=none,error bars/.cd,x dir=both,x explicit,error bar style={draw=black!80,line width=.65pt},error mark=-,error mark options={rotate=90,draw=black!80,mark size=1.8pt,line width=.65pt}] coordinates {(0.5018928538,3.25) += (0.1366143725,0) -= (0.1393367343,0) (0.4429664594,2.25) += (0.1335884589,0) -= (0.1389819566,0) (0.2853644369,1.25) += (0.0913816931,0) -= (0.0885614027,0) (0.0013577898,0.25) += (0.0008818578,0) -= (0.0006995749,0)};
\addplot[xbar,bar shift=0pt,bar width=5pt,fill=enc-paths,draw=black,line width=.35pt,mark=none,error bars/.cd,x dir=both,x explicit,error bar style={draw=black!80,line width=.65pt},error mark=-,error mark options={rotate=90,draw=black!80,mark size=1.8pt,line width=.65pt}] coordinates {(0.0815539598,3.00) += (0.0492014042,0) -= (0.0427386383,0) (0.0781163094,2.00) += (0.0489496936,0) -= (0.0355703282,0) (0.0556099500,1.00) += (0.0461183155,0) -= (0.0354410859,0) (0.0515056546,0.00) += (0.0136054246,0) -= (0.0123589820,0)};
\addplot[xbar,bar shift=0pt,bar width=5pt,fill=enc-local,draw=black,line width=.35pt,mark=none,error bars/.cd,x dir=both,x explicit,error bar style={draw=black!80,line width=.65pt},error mark=-,error mark options={rotate=90,draw=black!80,mark size=1.8pt,line width=.65pt}] coordinates {(0.5731077567,2.75) += (0.1292548823,0) -= (0.1369334846,0) (0.4769912967,1.75) += (0.1275969953,0) -= (0.1300213849,0) (0.3406208057,0.75) += (0.0990586779,0) -= (0.1003270596,0) (0.0936490451,-0.25) += (0.0180654040,0) -= (0.0182575795,0)};
\end{groupplot}\node[anchor=north,font=\fontsize{8}{9}\selectfont,align=center] at ([yshift=-9.5mm]group c2r1.south) {\tikz[baseline=-.3ex]{\path[fill=enc-dyadic,draw=black,line width=.3pt] (0,0) rectangle (1.2em,.65em);} Dyadic\quad\tikz[baseline=-.3ex]{\path[fill=enc-paths,draw=black,line width=.3pt] (0,0) rectangle (1.2em,.65em);} Paths\quad\tikz[baseline=-.3ex]{\path[fill=enc-local,draw=black,line width=.3pt] (0,0) rectangle (1.2em,.65em);} Local};
\end{tikzpicture}%
\endgroup}
\caption{\textbf{Reciprocal rank by observed endpoint connectivity.}
Panels (\textit{a})--(\textit{c}) report equal-family MRR across 57 knowledge graphs
for logistic regression, \textsc{CatBoost}, and \textsc{TabPFN-3}. Queries are
grouped as both directions, reverse only (candidate tail to head), direct only
(head to candidate tail), or no pair edge after removing the held-out target
edge. Within each group, bars report the \emph{Dyadic}, \emph{Paths}, and \emph{Local}
encoders from top to bottom. Error bars show 95\% provenance-family bootstrap sensitivity
intervals.}
\label{fig:query_evidence}
\end{figure}
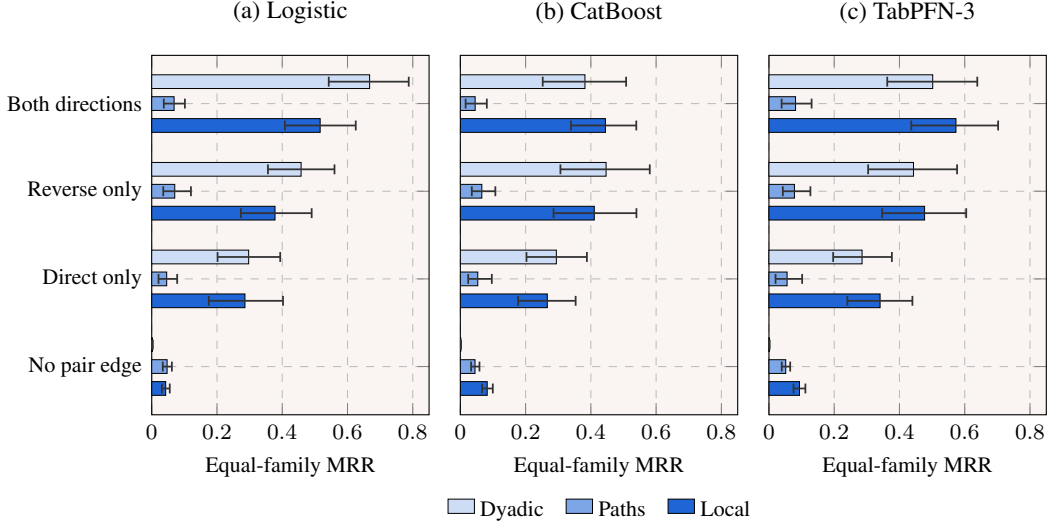

The \emph{Dyadic} encoder contains other-relation pair count, reverse-pair count, and a same-entity indicator. The \emph{Paths} encoder contains bounded two-hop and relation-conditioned path counts. The \emph{Local} encoder combines these groups with relation, endpoint-degree, and community features. \cref{fig:query_evidence} reports equal-family reciprocal rank within each stratum under the same evaluation queries, context, and seeds.

The connected strata contain 1,622 queries in total. \emph{Dyadic} features outperform \emph{Paths} when direct or reverse pair evidence is present. The no-pair-edge stratum contains \QueryNeither{} queries, for which \emph{Dyadic} reciprocal rank falls to \QueryDyadicNeitherRange{}, \emph{Paths} reaches \QueryPathsNeitherRange{}, and \emph{Local} reaches \QueryLocalNeitherRange{}. These fixed-panel comparisons identify where feature groups carry predictive evidence and remain descriptive at the feature-group level.

\section{Exploratory Analyses}
\label{app:screening}

The main analysis attributes performance through matched decoder comparisons. This section complements the paper's pre-specified contrasts with post-hoc analyses of shared difficulty and query heterogeneity. These analyses provide additional context for the interpretations in \cref{sec:feature-contributions} and \cref{sec:endpoint-connectivity}.

\subsection{Context-Based Rank Heuristic}

For numeric column \(j\), let \(n_+\) and \(n_-\) denote the numbers of positive and negative context rows. We compute its context separation as
\begin{equation}
A_j=\frac{1}{n_+n_-}\sum_{p=1}^{n_+}\sum_{q=1}^{n_-}\left[\mathbf{1}\!\left(x^+_{p,j}>x^-_{q,j}\right)+\frac{1}{2}\mathbf{1}\!\left(x^+_{p,j}=x^-_{q,j}\right)\right],
\label{eq:column-auc}
\end{equation}
where \(x^+_{p,j}\) and \(x^-_{q,j}\) are the values of column \(j\) for positive and negative context rows. We remove orientation by setting \(A_j^*=\max(A_j,1-A_j)\). With \(m\) the mean number of valid evaluation candidates per query, we define
\begin{equation}
\widehat R=(1-\max_j A_j^*)m.
\label{eq:screen_rank}
\end{equation}
\cref{eq:screen_rank} combines the strongest single-column separation observed in the labelled context with candidate-set size. It is a descriptive difficulty heuristic. It uses only labelled-context features and candidate-set size, excluding constant and nominal identifier columns.

\begin{figure}[t]
\centering
\includegraphics[width=0.8\linewidth]{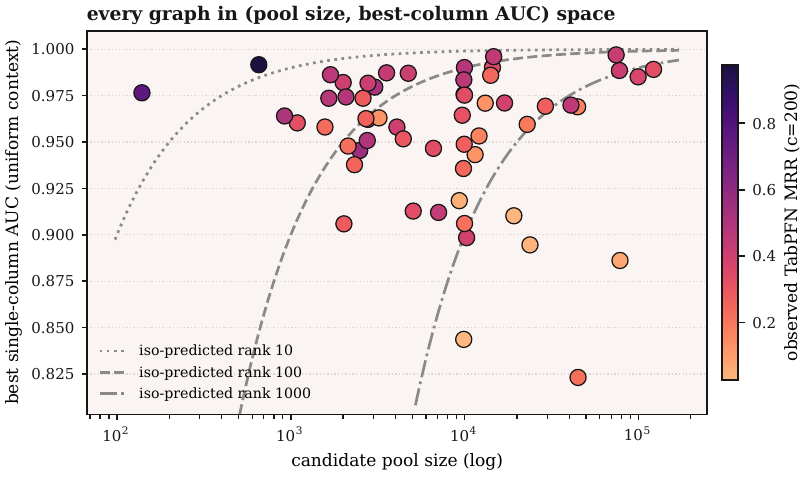}
\caption{\textbf{Context separability and candidate-pool size across the
57-KG panel.} Each point denotes one knowledge graph under the filtered-uniform
context with 200 examples per class. The horizontal axis reports candidate-pool
size on a logarithmic scale, the vertical axis reports the strongest oriented
single-column context ROC-AUC, and colour indicates observed
\textsc{TabPFN-3} MRR. Curves show equal values of the context-predicted rank.}
\label{fig:screen_anatomy}
\end{figure}

\cref{fig:screen_anatomy} shows the two quantities entering the heuristic. Graphs with similar single-column separation can differ in predicted difficulty because they rank different numbers of candidates, while stronger separation can offset a larger candidate pool.

\begin{figure}[t]
\centering
\includegraphics[width=\linewidth]{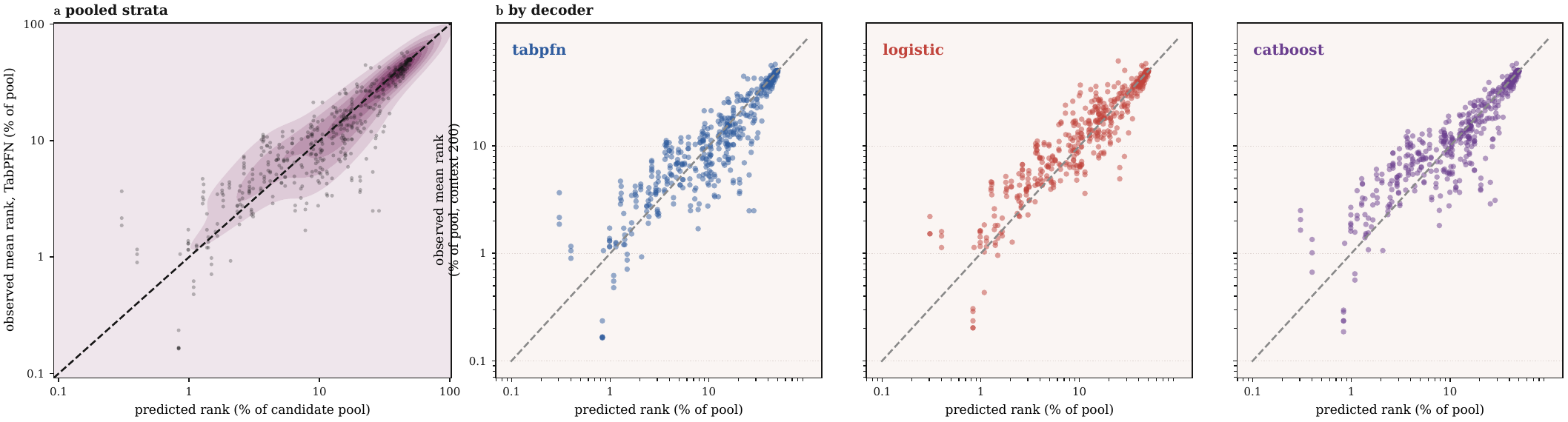}
\caption{\textbf{Transport of the context-predicted rank.} Panel (\textit{a}) compares predicted and observed mean rank across 455 dataset--encoder strata for \TabPFNthree{}; the solid curve and band summarise the observed strata. Panel (\textit{b}) shows the same predicted-versus-observed comparison separately for \TabPFNthree{}, logistic regression, and \CatBoost{}, left to right. Ranks are expressed as a percentage of the candidate pool on logarithmic axes, and dashed diagonals denote equality.}
\label{fig:rank_law_transport}
\end{figure}

Applied without refitting at 200 examples per class, \cref{eq:screen_rank} orders observed mean rank for \TabPFNthree{} at Spearman \(\ScreenRankSpearman{}\) across \(\ScreenStrata{}\) dataset--encoder strata. The corresponding correlations are 0.973 for logistic regression and 0.957 for \CatBoost{}. Of the 456 possible numeric dataset--encoder strata, \texttt{wiki\_mt2\_org} with the \emph{Dyadic} encoder is excluded because all three numerical columns are constant across all three seeds. \cref{fig:rank_law_transport} shows that the same envelope holds when the strata are separated by decoder.

The heuristic captures a substantial component of shared difficulty, while encoder and decoder selection remain separate questions. None of the \(\ScreenAssociationPairs{}\) benchmark-summary--configuration associations survives the joint max-\(|\rho|\) family-permutation sensitivity, and leave-one-family-out selectors do not improve on their in-fold majority baseline.

\subsection{Single-Column Context Separability}

\begin{figure}[!ht]
\centering
\includegraphics[width=\linewidth]{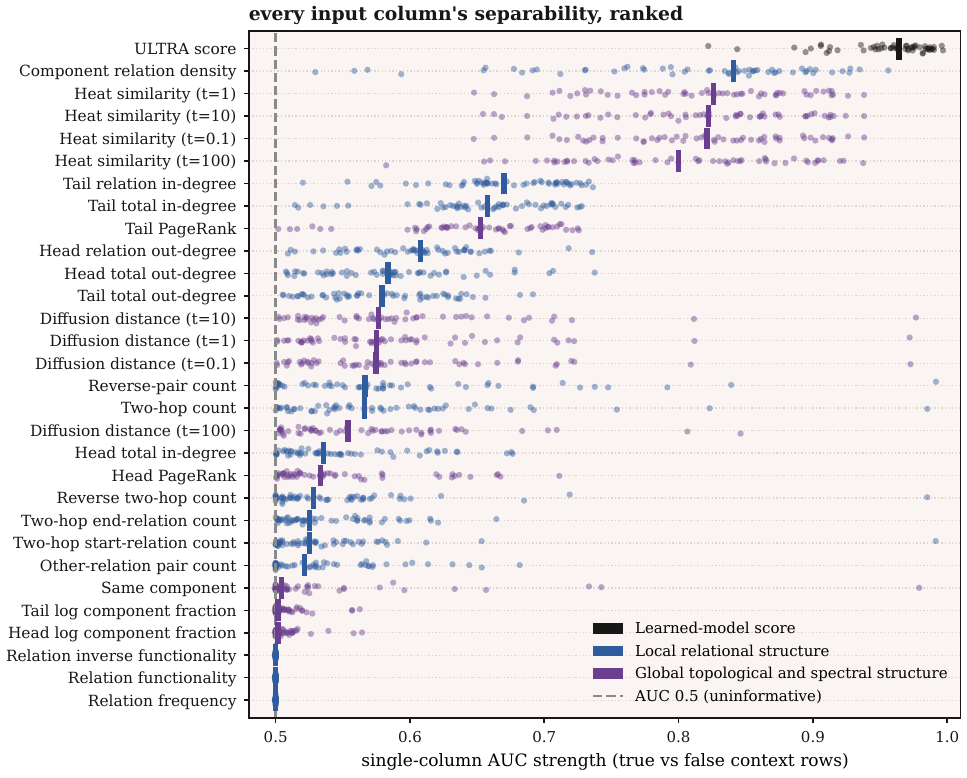}
\caption{\textbf{Single-column context separability across the 57-KG panel.}
Oriented ROC-AUC for every column of the \emph{Full} encoder at 200
examples per class. Each point denotes one knowledge graph and heavy ticks
denote column medians. Columns are ordered by median separability and grouped
into the frozen ULTRA score, local relational structure, and global topological
or spectral structure. The dashed reference marks chance separation at $0.5$.}
\label{fig:column_auc_spectrum}
\end{figure}

The column-level analysis in \cref{fig:column_auc_spectrum} follows the evidence-family taxonomy in \cref{tab:interfaces} and helps explain why the frozen score remains useful when added to local features. It is the strongest individual context separator on 55 of 57 graphs, whereas most constructed local and global columns lie closer to chance. This is a single-column diagnostic, while the main results concern multivariate and query-conditional effects beyond the strongest individual separator.

\subsection{Query-Level Variance and Structural Difficulty}

The endpoint-connectivity analysis in \cref{sec:endpoint-connectivity} conditions on graph evidence available before prediction. We additionally describe how much reciprocal-rank variation is associated with query identity and the remaining experimental factors. A sequential group-mean decomposition is computed in the declared order query, encoder, decoder, context, and seed. Median shares across datasets are \QueryVarianceShare{}, \InterfaceVarianceShare{}, \DecoderVarianceShare{}, 0.001, and less than 0.001, with 0.581 residual. Because the decomposition is sequential, these values are order-dependent descriptions.

The query--decoder interaction adds only \QueryDecoderInteractionShare{} median explained variance, with maximum 0.076. This provides limited evidence for per-query decoder selection under the present experimental setup. Within datasets, target-side relation degree correlates positively with reciprocal rank at \(\rho=\QueryTargetDegreeRho{}\), while anchor-side relation degree correlates negatively at \(\rho=\QueryAnchorDegreeRho{}\). MetaFam is the only dataset in which both directions reverse.

\begin{figure}[!ht]
\centering
\includegraphics[width=\linewidth]{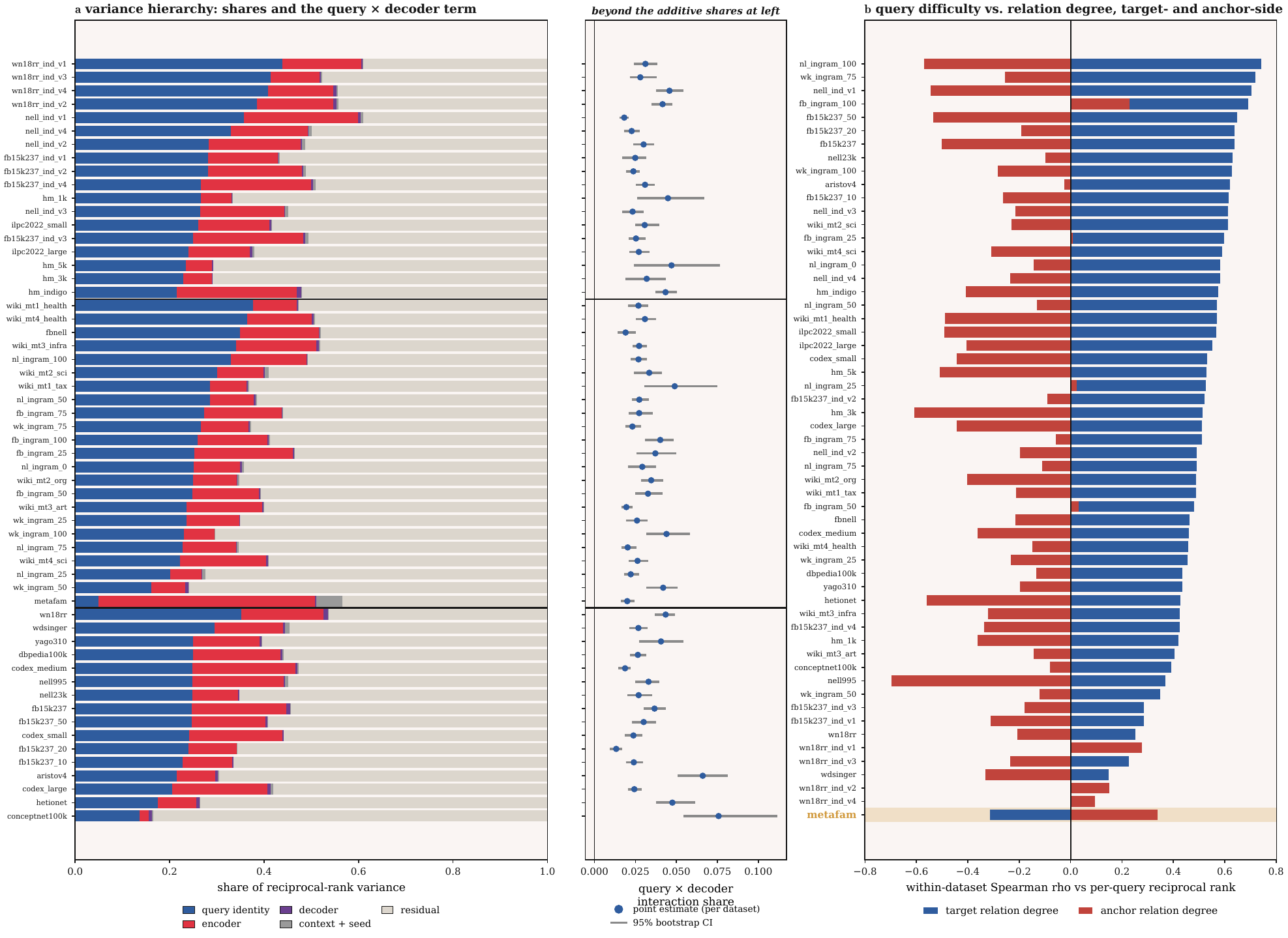}
\caption{\textbf{Query-level variance and structural difficulty across the 57-KG panel.} Panel (\textit{a}) reports the sequential reciprocal-rank shares assigned to query identity, encoder, decoder, context, and seed, together with the residual share and the additional query--decoder interaction.
The context and seed shares are combined only for display.
Panel (\textit{b}) reports within-dataset Spearman correlations between reciprocal rank and target- or anchor-side relation degree.
The decomposition in panel (\textit{a}) is descriptive and order-dependent.}
\label{fig:query_structure}
\end{figure}

\section{Audit Reporting Checklist}
\label{app:checklist}

\cref{tab:rules} condenses the comparisons used throughout the paper into a reporting checklist. It links each attribution question in \cref{sec:audit} to the failure mode that an end-to-end score would leave unresolved.

\begin{table}[!ht]
\centering
\scriptsize
\caption{\textbf{Reporting checklist for graph-to-table audits.}
Each row links an attribution question to the failure left unresolved by an
end-to-end score and the minimum matched comparison required to address it.}
\label{tab:rules}
\begin{adjustbox}{max width=\linewidth}
\begin{tabular}{p{0.18\linewidth}p{0.31\linewidth}p{0.35\linewidth}}
\toprule
Question & Failure to avoid & Required comparison \\
\midrule
Choose an encoder & Unique but non-reusable rows & Ceiling and singleton mass, with identifier relabelling where relevant \\
Add a feature group & Supported shortcut or nuisance information & Base, permuted, and aligned encoders \\
Choose a decoder & Decoder-specific encoder ranking & Same rows, context, and evaluation across decoders \\
Add an upstream score & Gain credited to the tabular readout & Nested encoders and native graph-model ranking \\
Interpret an average gain & Opposing query regimes hidden by averages & Paired query analysis grouped by observed graph evidence \\
\bottomrule
\end{tabular}
\end{adjustbox}
\end{table}

\section{Reproducibility}
\label{app:reproducibility}

\subsection{Decoder Configurations}

We report the default fixed decoder configurations used unless a decoder parameter is explicitly varied in a sensitivity analysis. Logistic regression uses \texttt{StandardScaler} fitted on context rows followed by \(C=1\) logistic regression. \CatBoost{} uses 500 iterations, depth 7, learning rate 0.05, and log loss. \TabPFNthree{} uses package version 8.1.0, the TabPFN-3 classifier checkpoint pinned by SHA-256 in the artefact, four estimators, \texttt{fit\_with\_cache}, and prediction batches capped at 50,000 rows. Automatic estimator scaling is disabled. A context-only sensitivity tunes logistic \(C\in\{0.01,0.1,1,10\}\) and \CatBoost{} depth \(\{4,7,10\}\) crossed with \(\ell_2\) leaf regularisation \(\{3,10\}\). Hyperparameter selection uses context data only.

\subsection{Categorical Encoding and Ranking}

We next document the decoder-specific transformations applied after the shared feature matrices are constructed. In the context-robustness experiment, relation ID and the same-partition indicator are treated as nominal for every decoder. Logistic regression uses one-hot encoding, \CatBoost{} uses native categorical features, and \TabPFNthree{} receives categorical indices. The same-partition value is a binary indicator of whether the two entities share a partition. Scaling and categorical transformations are fitted on context rows only.

Ranking uses the least-compressed positive-class score exposed by each decoder. For logistic regression with the one-column \Ultra{}-score encoder, the fitted coefficient determines orientation and the original upstream score is ranked directly, preserving native ties exactly. Every paired decoder comparison receives identical feature matrices, labels, context indices, and evaluation indices.

\subsection{Recorded Artefacts and Statistical Units}

We finally record the artefacts and replication units used to reproduce the reported estimates. The composition and 57-KG experiments record source revisions, model identities, seed streams, feature schemas, context and evaluation hashes, and prediction checksums. Reported uncertainty uses the replication unit declared in the corresponding experimental section, with graph, world, or provenance-family units.

\section*{GenAI Usage Statement}
We used generative AI tools to assist with manuscript revision, experiment planning, and implementation of additional analysis code. All AI-assisted code was reviewed and tested, numerical results were regenerated under frozen experimental plans, and AI-assisted text and claims were checked and revised by the authors. The authors take full responsibility for the final manuscript and accompanying artefacts.

\end{document}